\documentclass{article}

\PassOptionsToPackage{numbers, compress}{natbib}
\usepackage[preprint]{neurips_2023}

\usepackage[utf8]{inputenc} % allow utf-8 input
\usepackage[T1]{fontenc}    % use 8-bit T1 fonts
\usepackage{hyperref}       % hyperlinks
\usepackage{url}            % simple URL typesetting
\usepackage{booktabs}       % professional-quality tables
\usepackage{amsfonts}       % blackboard math symbols
\usepackage{nicefrac}       % compact symbols for 1/2, etc.
\usepackage{microtype}      % microtypography
\usepackage{xcolor}         % colors
\usepackage{enumitem}
\usepackage{graphicx}
\usepackage{scalerel}
\usepackage{multirow}
\usepackage{wrapfig}
\usepackage{subcaption}
\usepackage{bbm}

\usepackage{amsmath}
\usepackage{amssymb}
\usepackage{mathtools}
\usepackage{amsthm}

\theoremstyle{plain}
\newtheorem{theorem}{Theorem}[section]

\theoremstyle{definition}
\newtheorem{definition}[theorem]{Definition}
\newtheorem{assumption}[theorem]{Assumption}
\theoremstyle{remark}

\usepackage{hyperref}
\usepackage{algorithm}
\usepackage{algpseudocode}

\usepackage{tikz}
\usetikzlibrary{shapes,decorations,arrows,calc,arrows.meta,fit,positioning}
\tikzset{
    -Latex,auto,node distance =1 cm and 1 cm,semithick,
    state/.style ={ellipse, draw, minimum width = 0.7 cm},
    point/.style = {circle, draw, inner sep=0.04cm,fill,node contents={}},
    bidirected/.style={Latex-Latex,dashed},
    el/.style = {inner sep=2pt, align=left, sloped}
}

\newcommand\independent{\protect\mathpalette{\protect\independenT}{\perp}}
\def\independenT#1#2{\mathrel{\rlap{$#1#2$}\mkern2mu{#1#2}}}

\title{Causal Mediation Analysis with Multi-dimensional and Indirectly Observed Mediators}

\author{
  Ziyang Jiang$^1$ \quad Yiling Liu$^1$ \quad Michael H. Klein$^1$ \quad Ahmed Aloui$^1$ \quad Yiman Ren$^2$ \\
  \textbf{Keyu Li}$^1$ \quad \textbf{Vahid Tarokh}$^1$ \quad \textbf{David Carlson}$^1$ \\
  $^1$Duke University \quad $^2$University of Michigan Ross School of Business\\
  \texttt{\{ziyang.jiang,yiling.liu,michael.klein413,ahmed.aloui,keyu.li,} \\ \texttt{vahid.tarokh,david.carlson\}@duke.edu}\\
  \texttt{yimanren@umich.edu}
}

\begin{document}

\maketitle

\begin{abstract}
Causal mediation analysis (CMA) is a powerful method to dissect the total effect of a treatment into direct and mediated effects within the potential outcome framework.  This is important in many scientific applications to identify the underlying mechanisms of a treatment effect. However, in many scientific applications the mediator is unobserved, but there may exist related measurements.  For example, we may want to identify how changes in brain activity or structure mediate an antidepressant’s effect on behavior, but we may only have access to electrophysiological or imaging brain measurements. To date, most CMA methods assume that the mediator is one-dimensional and observable, which oversimplifies such real-world scenarios. To overcome this limitation, we introduce a CMA framework that can handle complex and indirectly observed mediators based on the identifiable variational autoencoder (iVAE) architecture. We prove that the true joint distribution over observed and latent variables is identifiable with the proposed method. Additionally, our framework captures a disentangled representation of the indirectly observed mediator and yields accurate estimation of the direct and mediated effects in synthetic and semi-synthetic experiments, providing evidence of its potential utility in real-world applications.

%Causal mediation analysis (CMA) is a powerful method for dissecting the total effect of a treatment into direct and indirect effects within the potential outcome framework. Unlike traditional mediation analysis, CMA does not require functional or distributional assumptions and is more adaptable to nonlinear causal relationships between variables. Nonetheless, most studies in the existing literature assume that the mediator is one-dimensional and observable, which oversimplifies real-world scenarios. To overcome this limitation, we introduce a CMA framework that can handle complex and unobserved mediators based on the model architecture of identifiable variational autoencoder (iVAE). We prove that the true joint distribution over observed and latent variables is identifiable with the proposed method. Additionally, our framework captures a disentangled representation of the unobserved mediator and yields accurate estimation of the direct and indirect effects across various experiments without observing the true mediator.
\end{abstract}

\section{Introduction}
\label{sec:1}
Causal inference methods are powerful tools to understand and quantify the causal relationships between treatments and outcomes, motivating studies in many areas \cite{athey2017state, glass2013causal,imai2010general,rothman2005causation}. Causal inference has been combined with machine learning in recent years to make powerful and flexible frameworks \cite{guo2020survey,li2022survey}. While these frameworks are highly useful to estimate the total treatment effect on an outcome, many scientific applications require understanding \emph{how} a treatment impacts outcomes.  This knowledge can then be used to design interventions that target the mediators to influence the outcome of interest. For example, we may want to identify neural changes that mediate a behavioral outcome when studying a treatment for a psychiatric disorder. Recent work has in fact found and \textit{manipulated} neural changes related to depression \cite{hultman2018brain} and social processing \cite{mague2022brain}.

This need motivates the usage of \textit{causal mediation analysis} (CMA), which estimates the causal effect on an outcome of interest that is due to changes in intermediate variables (the ``mediators’’) versus directly from the treatment \cite{pearl2001direct}. In specific contexts, understanding the role of the mediator is crucial as it tells us how nature works and provides insights into the underlying mechanisms that link variables, which enables a more accurate assessment of the treatment's effectiveness. In the above case, this means estimating how much of the behavior change is explained by the treatment’s impact on the brain, as well as how much behavioral change is unexplained by that relationship. Early studies on mediation analysis mainly adopted linear structural equation models (SEMs) including Wright's method of path analysis \cite{wright1923theory,wright1934method} and Baron and Kenny's method for testing mediation hypotheses \cite{baron1986moderator}. In the past few decades, researchers have come up with nonparametric generalizations for SEMs \cite{balke2013counterfactuals,joreskog1996nonlinear} which do not impose any functional or distributional forms on the causal relationships and therefore offer greater flexibility in modeling complex dependencies between variables.  

Despite these advances, a key challenge is that causal mediation analysis typically assumes a low-dimensional, often one-dimensional, mediator, whereas in many cases we want to identify mediation effects of complex data, such as neuroimaging, electrophysiology, and myriad -omics studies.  In this paper, we build upon the concept of the identifiable variational autoencoder (iVAE) \cite{khemakhem2020variational} and introduce a novel framework for CMA that can handle \emph{multi-dimensional} and \emph{indirectly observed} mediators.  We assume that there is a latent space that generates the high-dimensional observed data (e.g., a smaller latent space can generate the observed neural dynamics).  By using an identifiable model structure, we show that we can recover the latent space prior conditioned on the treatment and any available covariates.  In summary,  our main contributions are:
\begin{itemize}[leftmargin=*]
\item We propose a causal graph that involves both an \emph{indirectly observed} mediator and observed covariates that acts as a confounder for the treatment, the mediator, and the outcome.
\item We build a framework for CMA that can handle \emph{multi-dimensional} and \emph{indirectly observed} mediators based on the proposed causal graph.
\item We theoretically prove that the joint distribution over observed and latent variables in our framework is identifiable.
\item We show that our framework learns a disentangled representation of the \emph{indirectly observed} mediator between control and treatment groups. 
\item We empirically demonstrate the effectiveness of our framework on complex synthetic and semi-synthetic datasets.
\end{itemize}

\section{Related Work}
\label{sec:2}
\paragraph{Causal Mediation Analysis} 
As mentioned in the introduction, traditional mediation analysis was mainly based on linear SEMs where the direct, mediated, and total effects are determined by linear regression coefficients \cite{wright1923theory,wright1934method,baron1986moderator,mackinnon2012introduction,mackinnon1993estimating}.  Despite its simplicity, this approach relies on several assumptions such as normally distributed residuals \cite{pearl2014interpretation} and often leads to ambiguities when either the mediator or the outcome variable is not continuous \cite{rijnhart2019comparison}. To address this limitation, researchers formulated the causal mediation analysis (CMA) framework based on counterfactual thinking \cite{pearl2001direct,holland1988causal,rubin1974estimating}, which can accommodate nonlinear or nonparametric models such as targeted maximum likelihood estimation \cite{zheng2012targeted}, inverse propensity weighting (IPW) \cite{huber2013performance}, and natural effect models (NEMs) \cite{lange2012simple}. Within the counterfactual framework, the causal effects are calculated as the difference between two counterfactual combinations of mediators and outcomes, for which we will provide formal definitions in the next section. Although causal effects are defined at the individual level, in practice, we usually relax our estimation to their expected values over the population as we do not generally observe both potential outcomes simultaneously \cite{holland1986statistics}.

\paragraph{Causal Mediation Effect Estimation with Deep Models} 
Deep learning models have gained increasing attention for their capability in estimating causal effects within the potential outcome framework \cite{alaa2017bayesian, jiang2023estimating, louizos2017causal, shalit2017estimating}. In contrast, the use of deep learning models for mediation effect estimation has received comparatively less exploration. Xu et al. \cite{xu2022deepmed} developed a semiparametric neural network-based framework to reduce the bias in CMA. Cheng et al. \cite{cheng2022causal} and Xu et al. \cite{xu2023disentangled} used variational autoencoders (VAEs) to estimate the mediation effect based on a causal graph with hidden confounders. Although these VAE-based methods share some similarities with our proposed method, we distinguish ourselves by modeling the \emph{mediator} as the latent variable rather than the covariates, resulting in a different causal graph. Furthermore, these approaches assume that the mediator is observable and one-dimensional, which is not necessarily the case in many scientific applications.

\paragraph{Multi-dimensional Mediators} 
Compared to the many CMA methods proposed, significantly less research has been conducted on scenarios where the mediator is multi-dimensional and not directly observable. The majority of investigations on this subject are situated within the domains of neuroscience \cite{chen2018high,nath2023machine}, biostatistics \cite{zhang2021mediation}, and bioinformatics \cite{perera2022hima2,yang2021estimation,zhang2021survival,zhang2016estimating}. The approach proposed by Nath et al. \cite{nath2023machine} is the most relevant work to our research, where the high-dimensional mediator is first transformed into a one-dimensional variable, and the mediation effect is estimated using an iterative maximization algorithm. Nevertheless, all these methods primarily rely on linear SEMs and neglect the impact of any confounding variables, thereby limiting their applicability.

\section{Problem Setup}
\label{sec:3}

\begin{figure}[t!]
\centering
\begin{subfigure}[t]{0.47\textwidth}
    \centering
    \begin{tikzpicture}
    \node[state,circle,fill=black!15] (1) {$T$};
    \node[circle,draw,inner sep=4] (2) [above right =of 1, yshift=-0.5cm, xshift=0.3cm] {$Z$};
    \node[state,circle,fill=black!15] (3) [right =of 1, xshift=1.2cm, xshift=0.6cm] {$Y$};
    \node[state,circle,fill=black!15] (4) [above =of 2, yshift=-0.3cm] {$X$};
    \path (1) edge (2);
    \path (2) edge (3);
    \path (2) edge (4);
    \path (1) edge (3);
    \end{tikzpicture}
    \caption{}
    \label{fig:1a}
\end{subfigure}
\begin{subfigure}[t]{0.47\textwidth}
    \centering
    \begin{tikzpicture}
    \node[state,circle,fill=black!15] (1) {$T$};
    \node[circle,draw,inner sep=4] (2) [above right =of 1, yshift=-0.5cm, xshift=0.3cm] {$Z$};
    \node[state,circle,fill=black!15] (3) [right =of 1, xshift=1.2cm, xshift=0.6cm] {$Y$};
    \node[state,circle,fill=black!15] (4) [above =of 2, yshift=-0.3cm] {$X$};
    \node[circle,draw,fill=black!15,inner sep=3] (5) [below =of 2, yshift=-0.4cm] {$W$};
    
    \path (1) edge (2);
    \path (2) edge (3);
    \path (2) edge (4);
    \path (1) edge (3);
    \path (5) edge (1);
    \path (5) edge (2);
    \path (5) edge (3);
    \end{tikzpicture}
    \caption{}
    \label{fig:1b}
\end{subfigure}
\caption{Graphs of CMA for (a) case without observed covariates and (b) case with observed covariates, where $T$ is the treatment assignment, $Y$ is the outcome, $Z$ is the unobserved true mediator, $W$ is a set of observed covariates, and $X$ is a feature caused by the unobserved true mediator $Z$ with a much higher dimension. The observed variables are colored in grey.}
\label{fig:1}
\end{figure}
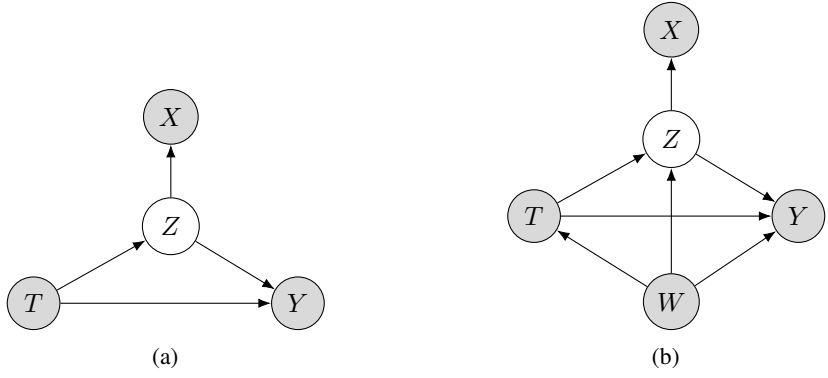

We assume that our causal model belongs to one of the two cases as displayed in Figure \ref{fig:1}. To ensure consistency with previous studies on mediation analysis \cite{pearl2014interpretation,hicks2011causal,mackinnon2007mediation}, we further assume that the treatment assignment $T$ is binary for each observed samples, with $T = 0$ indicating an assignment to the control group and $T = 1$ indicating an assignment to the treatment group. Consider the $n^{th}$ individual in an experiment with a total of $N$ units (i.e. $n = 1, ..., N$). Let $\boldsymbol{z}_n(t_n) \in \mathcal{Z} \subset \mathbb{R}^d$ denote the potential value of the unobserved true mediator under the treatment assignment $t_n$. Since $Y$ depends on both $T$ and $Z$, we denote $y(t_n, \boldsymbol{z}_n(t_n)) \in \mathcal{Y} \subset \mathbb{R}$ as the potential outcome of the $n^{th}$ individual under treatment $t_n$ and true mediator $\boldsymbol{z}_n(t_n)$. Following \cite{pearl2001direct,hicks2011causal,robins1992identifiability}, we can define the average causal mediation effects (ACME), the average direct effects (ADE), and the average total effect (ATE) as follows:
\begin{align}
ACME(t) &\coloneqq \mathbb{E} \left[ y(t, \boldsymbol{z}(1)) - y(t, \boldsymbol{z}(0)) \right], \label{eq:1} \\
ADE(t) &\coloneqq \mathbb{E} \left[ y(1, \boldsymbol{z}(t)) - y(0, \boldsymbol{z}(t)) \right], \label{eq:2} \\
ATE &\coloneqq \mathbb{E} \left[ y(1, \boldsymbol{z}(1)) - y(0, \boldsymbol{z}(0)) \right] \label{eq:3},
\end{align}
where the expectations are taken over all the samples in our experiment. Our main objective is to recover these quantities as accurately as possible. As $\boldsymbol{z}_n$ is unobserved, we must infer $\boldsymbol{z}_n$ from the related observed feature $\boldsymbol{x}_n \in \mathcal{X} \subset \mathbb{R}^D$ with a much higher dimension, i.e. $D \gg d$, as well as any other available information. In practice, there often exists a set of observed covariates $\boldsymbol{w}_n \in \mathcal{W} \subset \mathbb{R}^m$ that also acts as confounders for $T$, $Z$, and $Y$ as shown in Figure \ref{fig:1b}. With the presence of observed covariates, we need to make the following assumptions to make valid inferences about the causal effects:
\begin{assumption}
\label{assp:3.1}
There exists an observed variable $X \in \mathcal{X} \subset \mathbb{R}^{D}$ that is caused by the unobserved true mediator $Z \in \mathcal{Z} \subset \mathbb{R}^{d}$ as shown in Figure \ref{fig:1}.
\end{assumption}

\begin{assumption}
\label{assp:3.2}
The following two conditional independence assumptions hold sequentially.
\begin{align}
\left\{ Y(t',z), Z(t) \right\} &\independent T | W = w, \label{eq:4} \\
Y(t',z) &\independent Z(t) | T = t, W = w, \label{eq:5}
\end{align}
where $0 < p(T = t|W = w) < 1$, $0 < p(Z(t) = z|T = t, W = w) < 1$, and $t, t' \in \{0, 1\}$.
\end{assumption}

Assumption \ref{assp:3.2} is first introduced by Imai et al. \cite{imai2010general}, which is also known as \emph{sequential ignorability}. Note that Equation \ref{eq:4} is equivalent to the strong ignorability assumption common in causal inference \cite{rosenbaum1983central,rubin2005causal}. It states that the treatment assignment $T$ is statistically independent of potential outcome $Y$ and potential mediators $Z$ given covariates $W$. 
% This assumption is typically satisfied in randomized experiments and may also hold in observational studies if we collect as many pre-treatment covariates as possible, thereby making the ignorability of the treatment assignment more credible in such situations. 
Equation \ref{eq:5} states that given the treatment and covariates, the mediator $Z$ can be viewed as if it was randomized (in other words, there are no explained ``backdoor'' paths between the mediator and outcome \cite{pearl2014interpretation}).  

%implies that amongthose workers who share the same treatment status and the samepretreatment characteristics, the mediator can be regarded as if itwere randomized.
%Equation \ref{eq:5} states that the true mediator $Z$ is statistically independent of the potential outcome $Y($ given treatment assignment $T$ and covariates $W$. % This is a rather strong assumption that cannot be directly tested from the observed data, but its validity can be evaluated by conducting sensitivity analysis.

\section{Method}
\label{sec:4}
We leverage the model structure of identifiable variational autoencoder (iVAE) to estimate the causal mediation effects based on the causal graphs illustrated in Figure \ref{fig:1}. Our primary objective is to learn a disentangled representation of the true mediator in the latent space so that the statistical distance between $p(\boldsymbol{z}|t = 0)$ and $p(\boldsymbol{z}|t = 1)$ can be better estimated. In the following sections, we briefly review the concepts of identifiable variational autoencoder (iVAE) in Section \ref{sec:4.1}, present our framework in Section \ref{sec:4.2}, and formally state the identifiability of our framework in Section \ref{sec:4.3}.

\begin{wrapfigure}{R}{0.5\textwidth}
\centering
\vspace{-8mm}
\includegraphics[width=0.45\textwidth]{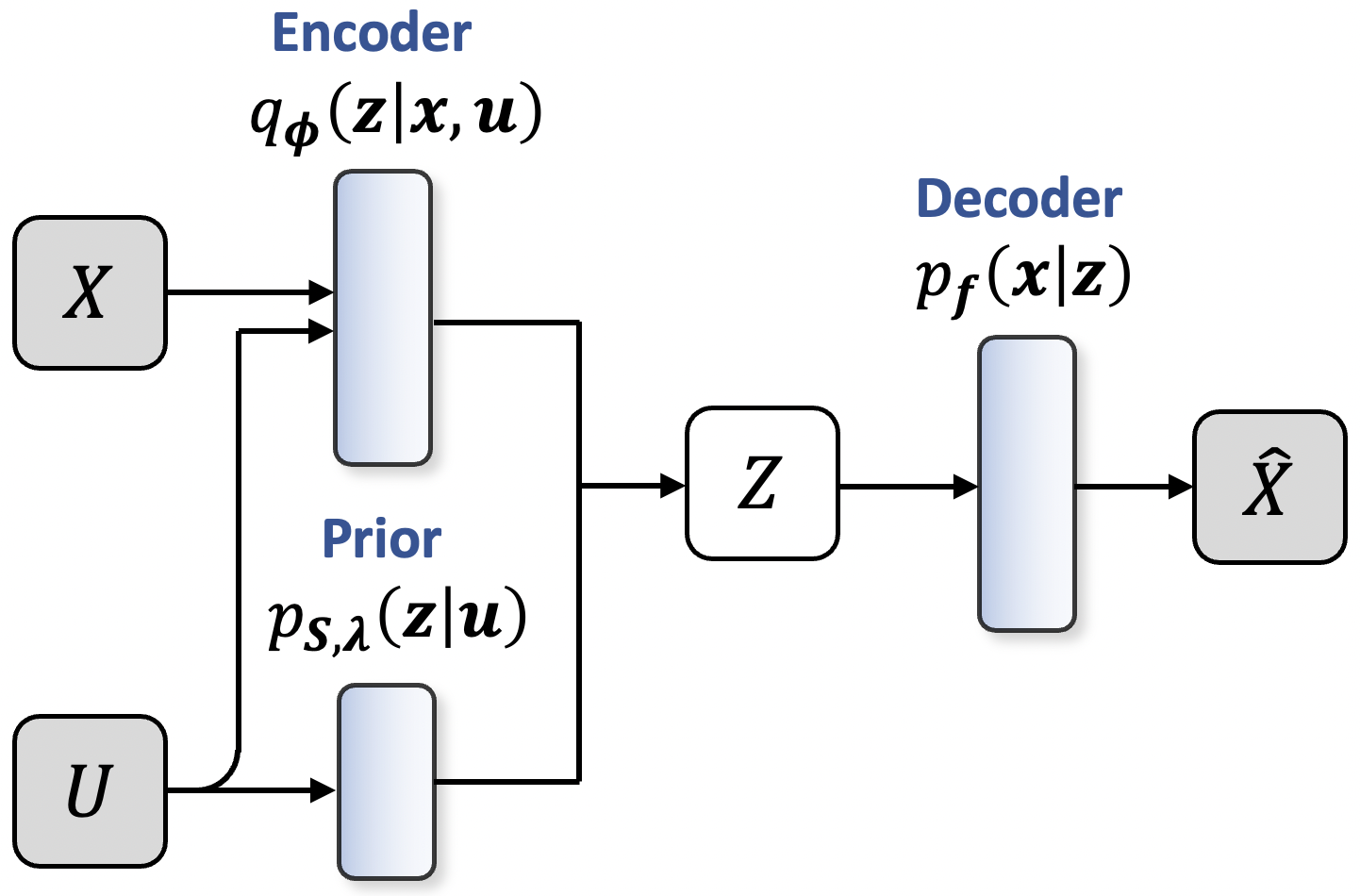}
\caption{Illustration of an iVAE where the blue nodes correspond to probabilistic distributions.}
\vspace{-4mm}
\label{fig:2}
\end{wrapfigure}

\subsection{Identifiable Variational Autoencoder (iVAE)}
\label{sec:4.1}
To begin with, here we provide a brief overview of iVAE  \cite{khemakhem2020variational}. We abuse the notation slightly by redefining $\boldsymbol{x}$ and $\boldsymbol{z}$ to refer to the observed data and the latent feature learned by a general variational autoencoder (VAE), respectively. The primary claim made by iVAE is that a VAE becomes identifiable up to \emph{a linear invertible transformation} (see Section \ref{sec:4.3} for full definitions) if we introduce a factorized prior distribution over the latent variable $\boldsymbol{z}$ conditioned on an auxiliary variable $\boldsymbol{u}$. Specifically, we have $\boldsymbol{z}$ 
sampled from $p(\boldsymbol{z}|\boldsymbol{u})$ which is assumed to be conditionally factorial with each $z_i \in \boldsymbol{z}$ belonging to a univariate exponential family as specified by the following probability density function:
\begin{equation}
p_{\boldsymbol{S},\boldsymbol{\lambda}}(\boldsymbol{z}|\boldsymbol{u}) = \prod_{i} \frac{Q_i(z_i)}{C_i(\boldsymbol{u})} \exp \left[ \sum_{j=1}^{k} S_{i,j}(z_i) \lambda_{i,j}(\boldsymbol{u}) \right], \label{eq:6}
\end{equation}
where $Q_i$ is the base measure, $C_i(\boldsymbol{u})$ is the normalizing constant, $k$ is a pre-defined number of sufficient statistics, $\boldsymbol{S}_i = (S_{i,1}, ..., S_{i,k})$ are the sufficient statistics, and $\boldsymbol{\lambda}_i(\boldsymbol{u}) = \left( \lambda_{i,1}(\boldsymbol{u}), ..., \lambda_{i,k}(\boldsymbol{u}) \right)$ are the natural parameters. 

The architecture of the iVAE framework is displayed in Figure \ref{fig:2}, which consists of a variational posterior $q_{\boldsymbol{\phi}}(\boldsymbol{z}|\boldsymbol{x}, \boldsymbol{u})$ and a conditional generative model $p_{\boldsymbol{\theta}}(\boldsymbol{x}, \boldsymbol{z}|\boldsymbol{u}) = p_{\textbf{f}}(\boldsymbol{x}|\boldsymbol{z}) p_{\boldsymbol{S},\boldsymbol{\lambda}}(\boldsymbol{z}|\boldsymbol{u})$ where $\textbf{f}$ is an injective function such that $p_{\textbf{f}}(\boldsymbol{x}|\boldsymbol{z}) = p_{\boldsymbol{\epsilon}}(\boldsymbol{x} - \textbf{f}(\boldsymbol{z}))$ and $\boldsymbol{\epsilon}$ is an independent noise variable with probability density function $p(\boldsymbol{\epsilon})$. The parameters of the generative model are denoted as $\boldsymbol{\theta} = \{\textbf{f}, \boldsymbol{S}, \boldsymbol{\lambda}\}$. When fitting iVAE on observed data, the parameter vector $(\boldsymbol{\theta}, \boldsymbol{\phi})$ is learned by maximizing the evidence lower bound (ELBO) $\mathcal{L}_{\boldsymbol{\theta}, \boldsymbol{\phi}}(\boldsymbol{x}, \boldsymbol{u})$:
\begin{equation}
\label{eqn:ELBO}
\log p_{\boldsymbol{\theta}}(\boldsymbol{x}|\boldsymbol{u}) \geq \mathcal{L}_{\boldsymbol{\theta}, \boldsymbol{\phi}}(\boldsymbol{x}, \boldsymbol{u}) \coloneqq \mathbb{E}_{q_{\boldsymbol{\phi}}(\boldsymbol{z}|\boldsymbol{x}, \boldsymbol{u})} \left[ \log p_{\boldsymbol{\theta}}(\boldsymbol{x}, \boldsymbol{z}|\boldsymbol{u}) - \log q_{\boldsymbol{\phi}}(\boldsymbol{z}|\boldsymbol{x}, \boldsymbol{u}) \right],
\end{equation}
where we use the reparameterization trick to sample from $q_{\boldsymbol{\phi}}(\boldsymbol{z}|\boldsymbol{x}, \boldsymbol{u})$. Briefly speaking, both the model structure and the learning process of iVAE are similar to conventional VAEs except that the prior, the variational posterior, and the decoder are additionally conditioned on the auxiliary variable $\boldsymbol{u}$. However, it is important to note that $\boldsymbol{u}$ must have some association with $\boldsymbol{x}$ and $\boldsymbol{z}$. %For example, if $\boldsymbol{x}$ represents a time series, $\boldsymbol{u}$ can be any relevant variable associated with the time series data, such as the corresponding time index, rather than an entirely unrelated variable.

\subsection{Estimating Mediation Effect with VAE}
\label{sec:4.2}

\begin{figure}[t!]
\centering
\begin{subfigure}{0.4\textwidth}
    \includegraphics[width=\linewidth]{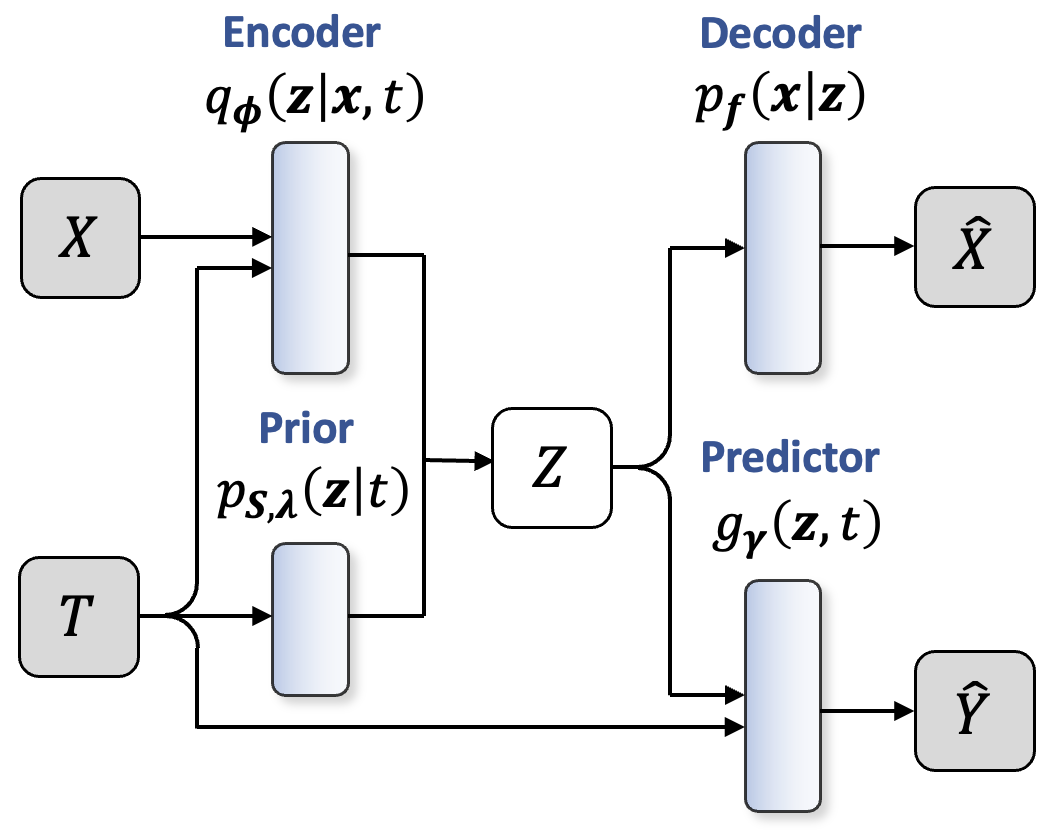}
    \caption{}
    \label{fig:3a}
\end{subfigure}
\hspace{15mm}
\begin{subfigure}{0.39\textwidth}
    \includegraphics[width=\linewidth,trim=2cm 0 0 0]{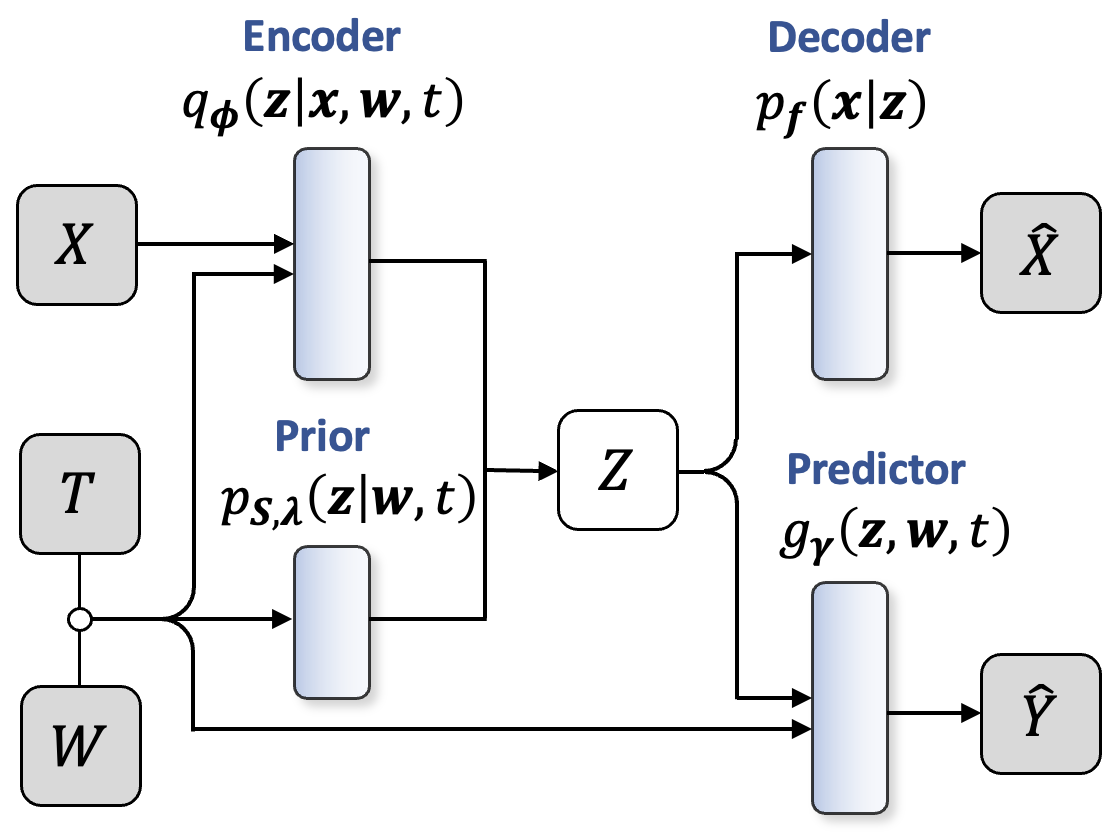}
    \caption{}
    \label{fig:3b}
\end{subfigure}
\caption{Illustration of the overall architecture of IMAVAE for (a) case without observed covariates and (b) case with observed covariates. Note that in case (b) the treatment assignment $T$ and the observed covariates $W$ are first concatenated and then passed into the prior, encoder, and decoder.}
\label{fig:3}
\end{figure}

In this section, we formally present our approach --- Identifiable Mediation Analysis with Variational Autoencoder (IMAVAE), with the overall architecture displayed in Figure \ref{fig:3}. The encoder, decoder, and prior components in IMAVAE have exactly the same probabilistic form as specified in Section \ref{sec:4.1} and share a similar structure with iVAE, where we take the high-dimensional feature $X$ as the input to the encoder to learn the unobserved mediator $Z$ and generate a reconstruction $\hat{X}$ with the decoder. Importantly, we further include a parametric model $g_{\boldsymbol{\gamma}}$ to predict the outcome $\hat{Y}$. Figures \ref{fig:3a} and \ref{fig:3b} depict two variants of our framework, corresponding to the two cases outlined in the causal graphs in Figure \ref{fig:1}:
\begin{itemize}[leftmargin=*]
\item \emph{Case (a)}: Without observed covariates, the treatment assignment $T$ is employed as the auxiliary variable and serves as input to the encoder, prior, and predictor, as illustrated in Figure \ref{fig:3a}.
\item \emph{Case (b)}: With observed covariates, we first concatenate the observed covariates $W$ and the treatment assignment $T$. The concatenated vector $(W,T)$ is then passed into the encoder, prior, and predictor as the auxiliary variable, as illustrated in Figure \ref{fig:3b}.
\end{itemize}

Similar to iVAE, we denote the parameter vector of IMAVAE as $(\boldsymbol{\theta}, \boldsymbol{\phi}, \boldsymbol{\gamma})$ where $\boldsymbol{\theta} = \{\textbf{f}, \boldsymbol{S}, \boldsymbol{\lambda}\}$. When fitting IMAVAE to the observed data, we optimize the parameter vector by minimizing the following objective:
\begin{equation}
\boldsymbol{\theta}^*, \boldsymbol{\phi}^*, \boldsymbol{\gamma}^* \coloneqq \arg \min_{\boldsymbol{\theta}, \boldsymbol{\phi}, \boldsymbol{\gamma}} \left\{ \alpha \mathcal{L}_{\boldsymbol{\theta}, \boldsymbol{\phi}} (\hat{\boldsymbol{x}}, \boldsymbol{x}) - \beta \mathcal{L}_{\boldsymbol{\theta}, \boldsymbol{\phi}}(\boldsymbol{x}, \boldsymbol{u}) + \mathcal{L}_{\boldsymbol{\phi}, \boldsymbol{S}, \boldsymbol{\lambda}, \boldsymbol{\gamma}}(\hat{y}, y) \right \}, \label{eq:8}
\end{equation}
where $\boldsymbol{u} = t$ for case (a), $\boldsymbol{u} = (\boldsymbol{w}, t)$ for case (b), $\mathcal{L}_{\boldsymbol{\theta}, \boldsymbol{\phi}} (\hat{\boldsymbol{x}}, \boldsymbol{x})$ is the discrepancy between the input feature $\boldsymbol{x}$ and its reconstruction $\hat{\boldsymbol{x}}$, $\mathcal{L}_{\boldsymbol{\phi}, \boldsymbol{S}, \boldsymbol{\lambda}, \boldsymbol{\gamma}}(\hat{y}, y)$ is the error between the predicted outcome $\hat{y}$ and the true outcome $y$. $\mathcal{L}_{\boldsymbol{\theta}, \boldsymbol{\phi}}(\boldsymbol{x}, \boldsymbol{u})$ represents the same loss term as Equation \ref{eqn:ELBO}. We note that this creates some overlap as the reconstruction term on $\boldsymbol{x}$ is also in Equation \ref{eqn:ELBO}, but choose this form to highlight each term independently (and does not change the overall loss with appropriately chosen weights). $\alpha$ and $\beta$ are hyperparameters representing the importance of the reconstruction error and the ELBO, respectively. In our experiments, we use mean squared error (MSE) loss for both $\mathcal{L}_{\boldsymbol{\theta}, \boldsymbol{\phi}} (\hat{\boldsymbol{x}}, \boldsymbol{x})$ and $\mathcal{L}_{\boldsymbol{\phi}, \boldsymbol{S}, \boldsymbol{\lambda}, \boldsymbol{\gamma}}(\hat{y}, y)$. The prior $p_{\boldsymbol{S},\boldsymbol{\lambda}}(\boldsymbol{z}|\boldsymbol{u})$ is set to be a multivariate normal distribution whose mean and covariance are parameterized as a function of $\boldsymbol{u}$ using a neural network.

To give an estimation on the direct, mediated, and total effects after fitting the parameters, we repeatedly sample $\boldsymbol{z}(t)$ from the learned distributions (i.e., $p_{\boldsymbol{S},\boldsymbol{\lambda}} (\boldsymbol{z}|t)$ for case (a) and $p_{\boldsymbol{S},\boldsymbol{\lambda}} (\boldsymbol{z}|
\boldsymbol{w}, t)$ for case (b)). Next, we feed both $\boldsymbol{z}(t)$ and the auxiliary variables into the predictor $g_{\boldsymbol{\gamma}}$ to obtain $y(t, \boldsymbol{z}(t))$ for case (a) or $y(t,\boldsymbol{w},\boldsymbol{z}(t))$ for case (b). Finally, we estimate the ACME, ADE, and ATE according to Equations \ref{eq:1}-\ref{eq:3} using estimated values of $y$.

\subsection{Identifiability of IMAVAE}
\label{sec:4.3}
In this section, we use similar definitions and assumptions stated by Khemakhem et al. \cite{khemakhem2020variational}. Specifically, let $\mathcal{Z} \subset \mathbb{R}^{d}$ be the support of distribution of $\boldsymbol{z}$. The support of distribution of $\boldsymbol{u}$ is $\mathcal{U} = \{0,1\}$ for case (a) and $\mathcal{U} = \{0,1\} \times \mathcal{W} \subset \mathbb{R}^{m+1}$ for case (b). We denote by $\textbf{S} \coloneqq (\textbf{S}_1, ..., \textbf{S}_d) = (S_{1,1}, ..., S_{d,k}) \in \mathbb{R}^{dk}$ the vector of sufficient statistics of Equation \ref{eq:6} and $\boldsymbol{\lambda}({\boldsymbol{u}}) = (\boldsymbol{\lambda}_1(\boldsymbol{u}), ..., \boldsymbol{\lambda}_d(\boldsymbol{u})) = (\lambda_{1,1}(\boldsymbol{u}), ..., \lambda_{d,k}(\boldsymbol{u})) \in \mathbb{R}^{dk}$ the vector of its parameters. Following the same notations in \cite{khemakhem2020variational}, we define $\mathcal{X} \subset \mathbb{R}^D$ as the image of \textbf{f} in Equation \ref{eq:6} and denote by $\textbf{f}^{-1}: \mathcal{X} \rightarrow \mathcal{Z}$ the inverse of $\textbf{f}$. Furthermore, we make the following assumption on the predictor:

\begin{assumption}
The predictor $g_{\boldsymbol{\gamma}}(\boldsymbol{z}, \boldsymbol{u})$ takes the following form:
\begin{equation}
g_{\boldsymbol{\gamma}}(\boldsymbol{z}, \boldsymbol{u}) \coloneqq p_{\textbf{h}}(y|\boldsymbol{z}, \boldsymbol{u}) = p_{\boldsymbol{\xi}}(y - \textbf{h}(\boldsymbol{z}, \boldsymbol{u})), \label{eq:9}
\end{equation}
where the function $\textbf{h}: \mathcal{Z} \times \mathcal{U} \rightarrow \mathcal{Y}$ is injective, $\mathcal{Y} \subset \mathbb{R}$ is the image of $\textbf{h}$, and $\boldsymbol{\xi}$ is an independent noise variable with probability density function $p_{\boldsymbol{\xi}}(\boldsymbol{\xi})$.
\end{assumption}

Similar to \cite{khemakhem2020variational}, for the sake of analysis, we treat $\textbf{h}$ as a parameter of the entire model and define $\boldsymbol{\psi} \coloneqq (\textbf{f}, \textbf{h}): \mathcal{Z} \times \mathcal{U} \rightarrow \mathcal{X} \times \mathcal{Y}$. $\boldsymbol{\psi}$ remains injective since both $\textbf{f}$ and $\textbf{h}$ are injective, and we consider the projection $\boldsymbol{\psi}^{-1}$ on $\mathcal{Z}$ to be $\boldsymbol{\psi}_{|\boldsymbol{z}}^{-1}$. The domain of parameters is thus $\Theta = \{\boldsymbol{\theta} \coloneqq (\textbf{f}, \textbf{h}, \textbf{S}, \boldsymbol{\lambda})\}$. To formally present our claim, we give the following definitions:
% With this, we define the domain of parameters as $\Theta = \Theta_1 \cup \Theta_2$ where $\Theta_1 = \{\boldsymbol{\theta} \coloneqq (\textbf{f}, \textbf{S}, \boldsymbol{\lambda})\}$ and $\Theta_2 = \{\boldsymbol{\theta} \coloneqq (\textbf{h}, \textbf{S}, \boldsymbol{\lambda})\}$. Since the identifiability of $(\textbf{f}, \textbf{S}, \boldsymbol{\lambda})$ is already proved by Khemakhem et al. \cite{khemakhem2020variational}, we focus on the identifiability of $(\textbf{h}, \textbf{S}, \boldsymbol{\lambda})$. To this end, we give the following definitions on identifiability:

%%% defintion of \psi
% $\boldsymbol{\psi} \coloneqq (\textbf{f}, \textbf{h}): \mathcal{Z} \times \mathcal{Z} \times \mathcal{U} \rightarrow \mathcal{X} \times \mathcal{Y}$. We consider the projection $\boldsymbol{\psi}^{-1}$ on $\mathcal{Z}\times \mathcal{Z}$ to be $\boldsymbol{\psi}_{|\boldsymbol{z}}^{-1}$.

%fixing the definition of psi

\begin{definition}
Let $\sim$ be an equivalence relation on $\Theta$. We say that $p_{\boldsymbol{\theta}}(\boldsymbol{x},\boldsymbol{z},y|\boldsymbol{u})$ is identifiable up to $\sim$ if $p_{\boldsymbol{\theta}}(\boldsymbol{x},\boldsymbol{z},y|\boldsymbol{u}) = p_{\boldsymbol{\tilde{\theta}}}(\boldsymbol{x},\boldsymbol{z},y|\boldsymbol{u}) \Longrightarrow \boldsymbol{\theta} \sim \boldsymbol{\tilde{\theta}}$.
\end{definition}

\begin{definition}
Let $\sim_{A}$ be the equivalence relation on $\Theta$ defined as follows:
\begin{equation}
(\textbf{f},\textbf{h},\textbf{S},\boldsymbol{\lambda}) \sim (\tilde{\textbf{f}},\tilde{\textbf{h}}, \tilde{\textbf{S}}, \tilde{\boldsymbol{\lambda}}) \Longleftrightarrow \exists A, \textbf{c} \,|\, \textbf{S}(\boldsymbol{\psi}^{-1}_{\boldsymbol{|z}}(\boldsymbol{x},y)) = A \tilde{\textbf{S}} (\tilde{\boldsymbol{\psi}}^{-1}_{|\boldsymbol{z}}(\boldsymbol{x},y)) + \textbf{c}, \forall \boldsymbol{x} \in \mathcal{X}; y \in \mathcal{Y}, 
\end{equation}
where $A$ is an invertible $dk \times dk$ matrix and $\textbf{c}$ is a vector.
\end{definition}

With all the assumptions and definitions stated above, we state our theorem below as an extension of the results in \cite{khemakhem2020variational}. The detailed proof will be provided in Appendix \ref{appx:A}.

\begin{theorem}\label{theo}
(Extension to Theorem 1 in Khemakhem et al. \cite{khemakhem2020variational}) Assume that we observe data sampled from the generative model $p_{\boldsymbol{\theta}}(\boldsymbol{x},\boldsymbol{z},y|\boldsymbol{u}) = p_{\textbf{f}}(\boldsymbol{x}|\boldsymbol{z})p_{\textbf{h}}(y|\boldsymbol{z}, \boldsymbol{u}) p_{\textbf{S},\boldsymbol{\lambda}} (\boldsymbol{z}|\boldsymbol{u})$ where $p_{\textbf{f}}(\boldsymbol{x}|\boldsymbol{z})$, $p_{\textbf{h}}(y|\boldsymbol{z}, \boldsymbol{u})$ and $p_{\textbf{S},\boldsymbol{\lambda}} (\boldsymbol{z}|\boldsymbol{u})$ follow the distributional form defined in Section \ref{sec:4.1}, Equation \ref{eq:9}, and Equation \ref{eq:6}, respectively. Then the parameters $(\textbf{f}, \textbf{h}, \textbf{S}, \boldsymbol{\lambda})$ will be $\sim_{A}$-identifiable if we assume the following holds:
\begin{enumerate}[leftmargin=*]
\item The set $\{(\boldsymbol{x}, y) \in \mathcal{X} \times \mathcal{Y} \,|\, \varphi_{\epsilon}(\boldsymbol{x}) = 0, \varphi_{\xi}(y) = 0\}$ has  measure zero, where 
$\varphi_{\boldsymbol{\epsilon}}$ and $\varphi_{\boldsymbol{\xi}}$ are the characteristic functions of $p_{\boldsymbol{\epsilon}}$ and $p_{\boldsymbol{\xi}}$ defined in Section \ref{sec:4.1} and Equation \ref{eq:9}, respectively. 
\item The functions $\textbf{f}$ and $\textbf{h}$ are both injective. 
\item The sufficient statistics $S_{i,j}$ in Equation \ref{eq:6} are differentiable almost everywhere, and $(S_{i,j})_{1\leq j \leq k}$ are linearly independent on any subset of $\mathcal{Y}$ of measure greater than zero.
\item There exists $dk+1$ distinct points $\boldsymbol{u}_0, ..., \boldsymbol{u}_{dk}$ such that the matrix $L = (\boldsymbol{\lambda}(\boldsymbol{u}_1) - \boldsymbol{\lambda}(\boldsymbol{u}_0), ..., \boldsymbol{\lambda}(\boldsymbol{u}_{dk}) - \boldsymbol{\lambda}(\boldsymbol{u}_0))$ of size $dk \times dk$ is invertible.
\end{enumerate}
\end{theorem}

With the aforementioned theorem, we state that the joint distribution learned by the generative model $p_{\boldsymbol{\theta}}(\boldsymbol{x},\boldsymbol{z},y|\boldsymbol{u})$ is identifiable.  Moreover, it is important to highlight that our extension of the identifiability theorem, originally presented in \cite{khemakhem2020variational}, incorporates the additional conditioning of $y$ on $\boldsymbol{u}$, thereby broadening the scope of iVAE.

\section{Experiments}
\label{sec:5}
We follow the approach using synthetic and semi-synthetic datasets used in recent causal inference manuscripts to allow for benchmarking and comparison of the results. We test IMAVAE on 3 datasets: 1 synthetic dataset and 2 semi-synthetic datasets\footnotemark{}. 
This allows us to evaluate how well we estimate counterfactual values of the treatment assignment and the mediator, and the direct, mediated, and total effects under reasonable assumptions. The detailed experimental setup (e.g. training details, computing resources, licenses, etc.) is given in Appendix \ref{appx:B}.
% \footnotetext[1]{Anonymous code is available at: \url{https://anonymous.4open.science/r/IMAVAE-557B/}.}
\footnotetext[1]{Code will be released upon acceptance.}

\begin{figure}[t!]
\centering
\hstretch{1.13}
{\includegraphics[scale=0.25]{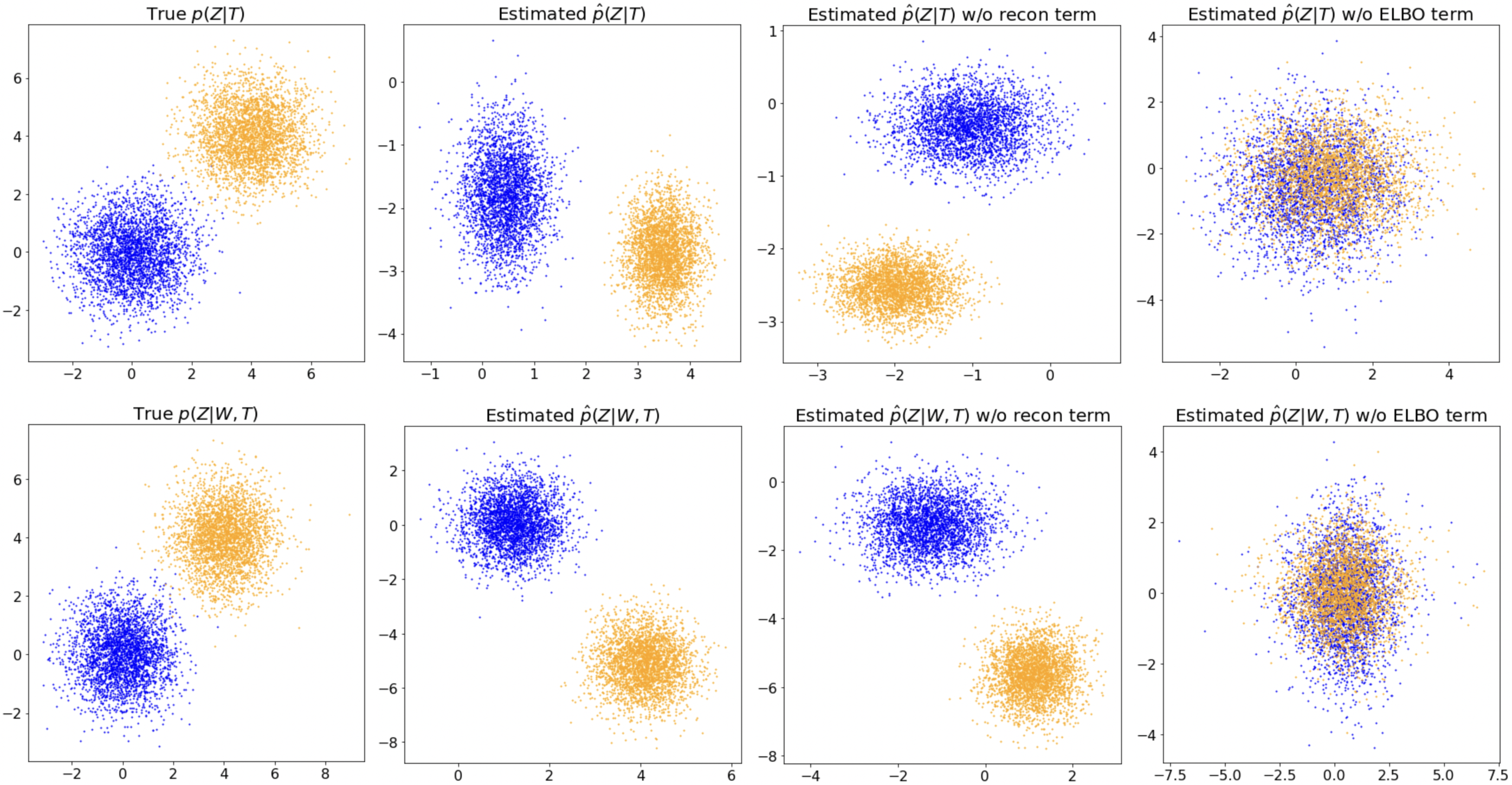}}
\caption{Distribution of the true and the estimated $p(\boldsymbol{z}|\boldsymbol{u})$ in the latent space where the upper row corresponds to case (a) without observed covariates, i.e. $\boldsymbol{u} = t$ and the bottom row corresponds for case (b) with observed covariates, i.e. $\boldsymbol{u} = (\boldsymbol{w}, t)$. From left to right, we present (left) the true distribution of $p(\boldsymbol{z}|\boldsymbol{u})$, (middle left) the estimated distribution $\hat{p}_{\boldsymbol{S},\boldsymbol{\lambda}} (\boldsymbol{z}|\boldsymbol{u})$ by IMAVAE, (middle right) the estimated distribution $\hat{p}_{\boldsymbol{S},\boldsymbol{\lambda}} (\boldsymbol{z}|\boldsymbol{u})$ without the reconstruction term, i.e. $\alpha = -1$, and (right) the estimated distribution $\hat{p}_{\boldsymbol{S},\boldsymbol{\lambda}} (\boldsymbol{z}|\boldsymbol{u})$ without the ELBO term, i.e. $\beta = 0$. The blue dots denote samples in control group and the orange dots denote samples in treatment group.}
\label{fig:4}
\end{figure}

\subsection{Synthetic Dataset}
\label{sec:5.1}

\begin{wraptable}{R}{0.61\textwidth}
\vspace{-7mm}
\begin{minipage}{0.61\textwidth}
\begin{table}[H]
\resizebox{\textwidth}{!}{
\begin{tabular}{c|ccc} 
\hline
             & IMAVAE & \begin{tabular}[c]{@{}c@{}}IMAVAE\\$\alpha = -1$\end{tabular} & \begin{tabular}[c]{@{}c@{}}IMAVAE\\$\beta = 0$\end{tabular}  \\ 
\hline
ACME ($t = 1$) &  0.056 $\pm$ .007      &  0.078 $\pm$ .007                                                                         &  2.875 $\pm$ .016                                                                         \\
ADE ($t = 0$)  &  0.058 $\pm$ .000      &  0.052 $\pm$ .000                                                                         &  0.043 $\pm$ .000                                                                         \\
ATE          &  0.003 $\pm$ .007      &  0.025 $\pm$ .006                                                                         &  2.917 $\pm$ .015                                                                         \\
\hline
\end{tabular}}
\vspace{2mm}
\caption{Absolute error of ACME under treated, ADE under control, and ATE on the synthetic dataset for IMAVAE in case (a) \emph{without} observed covariates}
\label{tab:1}
\end{table}
\vspace{-7mm}

\begin{table}[H]
\resizebox{\textwidth}{!}{
\begin{tabular}{c|ccc} 
\hline
             & IMAVAE & \begin{tabular}[c]{@{}c@{}}IMAVAE\\$\alpha = -1$\end{tabular} & \begin{tabular}[c]{@{}c@{}}IMAVAE\\$\beta = 0$\end{tabular}  \\ 
\hline
ACME ($t = 1$) &  0.214 $\pm$ .016      &  0.379 $\pm$ .014                                                                         &  5.912 $\pm$ .028                                                                         \\
ADE ($t = 0$)  &  0.194 $\pm$ .000      &  0.385 $\pm$ .000                                                                         &  0.127 $\pm$ .000                                                                         \\
ATE          &  0.019 $\pm$ .015      &  0.011 $\pm$ .016                                                                         &  6.036 $\pm$ .028                                                                         \\
\hline
\end{tabular}}
\vspace{2mm}
\caption{Absolute error of ACME under treated, ADE under control, and ATE on the synthetic dataset for IMAVAE in case (b) \emph{with} observed covariates}
\label{tab:2}
\end{table}
\end{minipage}
\vspace{-4mm}
\end{wraptable}

We first construct a synthetic dataset following the causal graphs in Figure \ref{fig:1}, where we give the details of data generation process in Appendix \ref{appx:C}. We set the unobserved true mediator to be two-dimensional (i.e. $d = 2$) for easier visualization. We display the distributions of the true and estimated unobserved mediator in Figure \ref{fig:4}, where we note that IMAVAE effectively learns \emph{disentangled representations} of $Z$ for the control and treatment groups in the latent space, up to trivial indeterminacies such as rotations and sign flips, for cases both with and without observed covariates. If we remove the reconstruction term (i.e. $\alpha = -1$ due to the overlap of reconstruction terms in Equation \ref{eq:8}), the shape and orientation of the distributions become slightly different but remain disentangled. However, if we discard the ELBO term (i.e. $\beta = 0$), the model fails to separate the distributions of control and treatment groups. We also compute the absolute errors between the estimated ACME, ADE, ATE, and their corresponding ground truths as shown in Tables \ref{tab:1} and \ref{tab:2}. It can be observed that IMAVAE yields slightly larger errors when the reconstruction term is removed (i.e., $\alpha = -1$). However, without the ELBO term (i.e., $\beta = 0$), the model produces significantly larger errors on ACME and ATE. From the obtained results, we conclude that the iVAE-like structure in our framework is essential for learning a better representation of the unobserved mediator, which, in turn, improves the accuracy of mediation effect estimation.

\subsection{Electrophysiological Dataset}
\label{sec:5.2}

As described in the introduction, causal mediation analysis holds significant relevance for applications in systems neuroscience. 
%Many works leverage machine learning techniques to learn explainable latent variables correlated to a behavior of interest from high dimensional time-series data \cite{}. For example, some recent works have made use of multi-site brain recordings to identify brain-networks associated with social activity \cite{}, stress \cite{}, and depression susceptibility \cite{} with the aim of developing better treatments for psychiatric illnesses. 
Once such area is in the emerging area of targeted neurostimulation (see \cite{Deisseroth2011,Limousin2019,Tufail2011} for a description), where the brain is manipulated by optical, electrical, or mechanical stimulation with the goal of manipulating behavior in many brain conditions.  However, while the mechanism of behavioral change is the brain, identifying such changes is challenging with existing causal mediation techniques due to the high-dimensional and complex nature of the brain data. Accurate appraisal of neural changes causing behavioral change will provide a deeper understanding of mechanisms driving neural activity and potentially lead to more efficacious treatments. 

%Many current methods for causal mediation analysis are ill-suited to this domain of problems as these latent variables are not observed directly but are inferred from the observed neural activity. Additionally, it is often beneficial to consider latent variables with more than one dimension to study their interactions, or to capture more nuanced behaviors or activity. Our method overcomes these shortcomings and provides a means to evaluate the causal effects of multiple mediators. 

We demonstrate capability of our method to this domain with a semisynthetic dataset by post-processing real multi-site brain recordings using local field potential (LFP) data from 26 mice \cite{gallagher2017cross}, which is publicly available \cite{carlson2023data}. Each mouse is recorded by a certain number of time steps, resulting in a total of 43,882 data points. We take the LFP signals as the observed feature $X$, while the true mediator $Z$ is manually generated by applying principal component analysis (PCA) to map $X$ into a lower-dimensional representation. The treatment assignment $T$ indicates whether the mouse is recorded during an open field exploration $(T = 0)$ or a tail suspension test $(T = 1)$. Furthermore, we consider the genotype of the mouse, a binary variable, as an observed covariate, denoted by $W \in {0, 1}$. Lastly, we construct the outcome $Y$ manually as a function of the treatment, the mediator, and the genotype (only for case (b) with observed covariate). The detailed procedure of dataset generation is given in Appendix \ref{appx:D}.

We compare our method with two baseline models that are designed to handle high-dimensional mediators: an integrated framework of shallow or deep neural network and linear SEM (Shallow/Deep LSEM) \cite{nath2023machine} and a high-dimensional mediation analysis (HIMA) framework \cite{zhang2016estimating}. Notably, HIMA considers each component of $Z$ as an individual mediator instead of a multidimensional mediator. As such, we report the mediation effect using the component with the highest correlation. We compute and display the absolute errors of ACME, ADE, and ATE in Table \ref{tab:3}. Our results indicate that IMAVAE outperforms both benchmarks by a very weide margin on all estimations except the ATE in case (a) without covariates. The two benchmarks used in this experiment yield significantly larger errors on ACME and ADE. We believe this is reasonable, as both benchmarks are designed based on linear SEMs and are thus not able to capture the correlation between the components of $Z$.

\begin{table}[t!]
\setlength{\tabcolsep}{0.45em}
\centering
\caption{Absolute error of ACME under treated, ADE under control, and ATE on the tail suspension test dataset for IMAVAE and other benchmarks.}
\vspace{2mm}
\resizebox{\textwidth}{!}{
\begin{tabular}{c|cccc|cccc} 
\hline
             & \multicolumn{4}{c|}{Case (a)}                                                                                                                                              & \multicolumn{4}{c}{Case (b)}                                                                                                                                                \\
             & \begin{tabular}[c]{@{}c@{}}IMAVAE\\(ours)\end{tabular} & \begin{tabular}[c]{@{}c@{}}Shallow\\LSEM\end{tabular} & \begin{tabular}[c]{@{}c@{}}Deep\\LSEM\end{tabular} & HIMA & \begin{tabular}[c]{@{}c@{}}IMAVAE\\(ours)\end{tabular} & \begin{tabular}[c]{@{}c@{}}Shallow\\LSEM\end{tabular} & \begin{tabular}[c]{@{}c@{}}Deep\\LSEM\end{tabular} & HIMA  \\ 
\hline
\begin{tabular}[c]{@{}c@{}}ACME\\$(t = 1)$\end{tabular} &  \textbf{2.348 $\pm$ 0.003}                                                      &  15.14 $\pm$ 0.03                                                     &  15.48 $\pm$ 0.07                                                  &  13.95 $\pm$ 0.03    &  \textbf{0.559 $\pm$ 0.002}                                                      &  3.06 $\pm$ 2.16                                                     &  4.80 $\pm$ 0.44                                                  &  2.67 $\pm$ 0.02    \\
\begin{tabular}[c]{@{}c@{}}ADE\\$(t = 0)$\end{tabular}  &  \textbf{1.603 $\pm$ 0.000}                                                      &  14.71 $\pm$ 0.03                                                     &  15.06 $\pm$ 0.07                                                  &  3.82 $\pm$ 0.01   &  \textbf{0.782 $\pm$ 0.000}                                                      &  5.03 $\pm$ 1.20                                                     &  5.16 $\pm$ 0.51                                                  &  3.21 $\pm$ 0.01     \\
ATE          &  0.744 $\pm$ 0.003                                                      &  \textbf{0.42 $\pm$ 0.06}                                                     &  0.42 $\pm$ 0.14                                                  &  17.77 $\pm$ 0.03    &  \textbf{0.223 $\pm$ 0.002}                                                      &  1.96 $\pm$ 3.36                                                     &  0.36 $\pm$ 0.94                                                  &  0.54 $\pm$ 0.02      \\
\hline
\end{tabular}}
\label{tab:3}
\end{table}

\subsection{Jobs II Dataset}
\label{sec:5.3}

To evaluate whether our method can generalize to real-world scenarios used in recent causal mediation analysis frameworks, we test IMAVAE on the Jobs II dataset \cite{vinokur1995impact}, which aims to explore the impact of unemployment on workers' stress and mental health and evaluate the potential benefits of participation in a job-search skills seminar. The dataset includes a binary treatment assignment $T$, which indicates whether a participant was assigned to attend a job-search skills seminar ($T = 1$) or to receive a booklet ($T = 0$). The mediator $Z$ is a continuous variable that measures the job-search efficacy. All other attributes are treated as the observed covariates $W$. The outcome variable $Y$ is also a continuous variable that represents the level of depression reported by each participant during follow-up interviews. To obtain the ground truth for direct and mediated effects, we followed a simulation procedure similar to \cite{huber2016finite} to make ACMEs, ADEs, and ATE all equal to zero. The detailed simulation procedure is given in Appendix \ref{appx:E}. 

We compare the performance of our method with several benchmarks: nonlinear SEM with interaction (LSEM-I) \cite{imai2010general}, imputing-based natural effect model (NEM-I) \cite{lange2012simple}, IPW \cite{huber2013performance}, and Causal Mediation Analysis with Variational Autoencoder (CMAVAE) \cite{cheng2022causal}. It is worth noting that the Jobs II dataset presents an observable mediator variable $Z$, which is \emph{not} the optimal scenario for our proposed framework, as IMAVAE is specifically designed for CMA with \emph{implicitly} observed mediators. Nonetheless, according to the results shown in Tables \ref{tab:4} and \ref{tab:5} (where $N$ is the total number of simulated samples and $\eta$ is a simulation parameter which stands for the magnitude of selection into the mediator), our method still mostly outperforms the benchmarks in terms of the estimation on ACME, ADE, and ATE with a reasonable level of uncertainty.

\section{Discussion}
\label{sec:6}

\paragraph{Design Choice of the Predictor} 
In Section \ref{sec:4.3}, we prove that the true joint distribution over observed and latent variables learned by IMAVAE is identifiable if we specify the predictor $g_{\boldsymbol{\gamma}}$ to be a conditional distribution reparameterized by function $\textbf{h}$. However, in practice, $g_{\boldsymbol{\gamma}}$ can be as simple as a linear or logistic regression model since we believe the identifiability on $(\textbf{f}, \textbf{S}, \boldsymbol{\lambda})$ is enough to disentangle the representations between control and treatment groups and give an accurate estimation on the mediation effects. We encourage readers to consider designing $g_{\boldsymbol{\gamma}}$ in order to achieve optimal performance.

\paragraph{Limitations}
As discussed in Section \ref{sec:3}, our method relies on sequential ignorability, a condition that is not directly testable using the observed data. However, recent studies \cite{cheng2022causal, xu2023disentangled} propose a potential solution by considering $X$ as a proxy variable and accounting for hidden confounders. Exploring this approach represents an intriguing direction for our future research.

\paragraph{Applications and Broader Impacts}
We believe the proposed model architecture can be very useful for improving interpretability for neuroscience applications. For instance, the disentangled mediator representations obtained by IMAVAE can be used to investigate the brain activities of individuals under different interventions. It can also be combined with other interpretable methods such as linear factor models to better illustrate the high-dimensional dynamics in brain networks as proposed by Talbot et al \cite{talbot2020supervised}. We have not identified any potential negative societal consequences specific to this manuscript.

% \paragraph{Interpretability?}
% \textcolor{red}{Adapting this model architecture to improve interpretability for neuroscience applications would be beneficial and is held for future work (Hunter)}

\begin{table}[t!]
\setlength{\tabcolsep}{0.45em}
\centering
\caption{Absolute error of ACME under treated, ADE under control, and ATE on simulated Jobs II data for IMAVAE and other benchmarks where 10\% of the data are mediated (i.e. $Z > 3$).}
\vspace{2mm}
\resizebox{\textwidth}{!}{
\begin{tabular}{c|cccccccccc} 
\hline
         & \multicolumn{2}{c}{LSEM-I} & \multicolumn{2}{c}{NEM-I} & \multicolumn{2}{c}{IPW} & \multicolumn{2}{c}{CMAVAE} & \multicolumn{2}{c}{IMAVAE (ours)}  \\
$N$       & 500 & 1000                & 500 & 1000                & 500 & 1000              & 500 & 1000                 & 500 & 1000                         \\
\hline
         & \multicolumn{10}{c}{ACME under treated $(t = 1)$}                                                                                                                                                                   \\ 
\hline
$\eta = 10$ &  0.9 $\pm$ .04   &  0.6 $\pm$ .02                  &  0.6 $\pm$ .03   &  0.8 $\pm$ .01                   &  0.6 $\pm$ .04   &  0.8 $\pm$ .02                 &  0.2 $\pm$ .00   &  0.3 $\pm$ .00                    &  \textbf{0.1 $\pm$ .02}   &  \textbf{0.1 $\pm$ .01}                            \\
$\eta = 1$  &  0.0 $\pm$ .01   &  0.1 $\pm$ .01                    &  \textbf{0.0 $\pm$ .00}   &  0.1 $\pm$ .01                   &  0.0 $\pm$ .01   &  0.1 $\pm$ .01                 &  0.1 $\pm$ .00   &  0.1 $\pm$ .00                    &  0.1 $\pm$ .01   &  \textbf{0.0 $\pm$ .01}                            \\ 
\hline
         & \multicolumn{10}{c}{ADE under control $(t = 0)$}                                                                                                                                                                    \\ 
\hline
$\eta = 10$ &  1.3 $\pm$ .07   &  1.6 $\pm$ .06                    &  1.2 $\pm$ .06   &  1.8 $\pm$ .05                   &  1.2 $\pm$ .06   &  0.2 $\pm$ .06                 &  \textbf{0.1 $\pm$ .00}   &  \textbf{0.0 $\pm$ .03}                    &  0.3 $\pm$ .00   &  0.3 $\pm$ .00                            \\
$\eta = 1$  &  3.3 $\pm$ .08   &  \textbf{0.0 $\pm$ .07}                    &  1.1 $\pm$ .03  &  0.2 $\pm$ .07   &  3.3 $\pm$ .08                   &  0.3 $\pm$ .06   &  0.5 $\pm$ .02                 &  0.4 $\pm$ .01   &  \textbf{0.2 $\pm$ .00}                    &  0.3 $\pm$ .00                          \\ 
\hline
         & \multicolumn{10}{c}{ATE}                                                                                                                                                                                  \\ 
\hline
$\eta = 10$ &  2.2 $\pm$ .05   &  1.0 $\pm$ .06                    &  1.8 $\pm$ .05   &  0.9 $\pm$ .06                   &  0.5 $\pm$ .05   &  1.0 $\pm$ .06                 &  0.3 $\pm$ .01   &  0.3 $\pm$ .03                    &  \textbf{0.2 $\pm$ .02}   &  \textbf{0.2 $\pm$ .01}                            \\
$\eta = 1$  &  3.3 $\pm$ .08   &  0.1 $\pm$ .07                    &  3.4 $\pm$ .03   &  \textbf{0.1 $\pm$ .06}                   &  3.2 $\pm$ .07   &  0.2 $\pm$ .05                 &  0.4 $\pm$ .02   &  0.3 $\pm$ .01                    &  \textbf{0.2 $\pm$ .01}   &  0.2 $\pm$ .01                            \\
\hline
\end{tabular}}
\label{tab:4}
\end{table}

\begin{table}[t!]
\setlength{\tabcolsep}{0.45em}
\centering
\caption{Absolute error of ACME under treated, ADE under control, and ATE on simulated Jobs II data for IMAVAE and other benchmarks where 50\% of the data are mediated (i.e. $Z > 3$)}
\vspace{2mm}
\resizebox{\textwidth}{!}{
\begin{tabular}{c|cccccccccc} 
\hline
         & \multicolumn{2}{c}{LSEM-I} & \multicolumn{2}{c}{NEM-I} & \multicolumn{2}{c}{IPW} & \multicolumn{2}{c}{CMAVAE} & \multicolumn{2}{c}{IMAVAE (ours)}  \\
$N$       & 500 & 1000                & 500 & 1000                & 500 & 1000              & 500 & 1000                 & 500 & 1000                         \\
\hline
         & \multicolumn{10}{c}{ACME under treated $(t = 1)$}                                                                                                                                                                   \\ 
\hline
$\eta = 10$ &  0.9 $\pm$ .03   &  0.6 $\pm$ .03                  &  0.2 $\pm$ .03   &  0.4 $\pm$ .03                   &  0.2 $\pm$ .03   &  0.4 $\pm$ .03                 &  \textbf{0.0 $\pm$ .00}   &  0.1 $\pm$ .00                    &  0.0 $\pm$ .05   &  \textbf{0.0 $\pm$ .03}                            \\
$\eta = 1$  &  0.1 $\pm$ .01   &  \textbf{0.0 $\pm$ .01}                    &  0.2 $\pm$ .00   &  0.1 $\pm$ .01                  &  0.1 $\pm$ .01   &  \textbf{0.0 $\pm$ .01}                 &  0.1 $\pm$ .00   &  0.1 $\pm$ .00                    &  \textbf{0.0 $\pm$ .05}   &  0.0 $\pm$ .03                            \\ 
\hline
         & \multicolumn{10}{c}{ADE under control $(t = 0)$}                                                                                                                                                                    \\ 
\hline
$\eta = 10$ &  0.6 $\pm$ .06   &  0.1 $\pm$ .04                    & 0.1 $\pm$ .06   &  0.1 $\pm$ .04                   &  0.7 $\pm$ .07   &  0.2 $\pm$ .05                 &  0.3 $\pm$ .01   &  \textbf{0.1 $\pm$ .00}                   &  \textbf{0.1 $\pm$ .00}   &  \textbf{0.1 $\pm$ .00}                            \\
$\eta = 1$  &  0.1 $\pm$ .10   &  0.3 $\pm$ .10                    &  0.1 $\pm$ .10  &  0.3 $\pm$ .04   &  0.3 $\pm$ .10                   &  0.2 $\pm$ .04   &  \textbf{0.1 $\pm$ .00}                 &  \textbf{0.1 $\pm$ .00}   &  \textbf{0.1 $\pm$ .00}                    &  \textbf{0.1 $\pm$ .00}                          \\ 
\hline
         & \multicolumn{10}{c}{ATE}                                                                                                                                                                                  \\ 
\hline
$\eta = 10$ &  0.3 $\pm$ .05   &  0.8 $\pm$ .03                    &  0.1 $\pm$ .05   &  0.5 $\pm$ .03                   &  0.9 $\pm$ .05   &  0.2 $\pm$ .04                 &  0.3 $\pm$ .01   &  \textbf{0.0 $\pm$ .01}                    &  \textbf{0.1 $\pm$ .04}   &  0.1 $\pm$ .03                            \\
$\eta = 1$  &  0.1 $\pm$ .09   &  0.3 $\pm$ .04                    &  0.3 $\pm$ .10   &  0.3 $\pm$ .04                   &  0.2 $\pm$ .10   &  0.2 $\pm$ .04                 &  \textbf{0.0 $\pm$ .01}   &  0.2 $\pm$ .01                    &  0.1 $\pm$ .04   &  \textbf{0.1 $\pm$ .03}                            \\
\hline
\end{tabular}}
\label{tab:5}
\end{table}

\section{Conclusion}
\label{sec:7}
This work makes a contribution to the field of causal mediation analysis (CMA) by proposing a novel method, IMAVAE, that can handle situations where the mediator is indirectly observed and observed covariates are likely to be present. Our approach builds on existing CMA methods and leverages the identifiable variational autoencoder (iVAE) model architecture to provide a powerful tool for estimating direct and mediated effects. We have demonstrated the effectiveness of IMAVAE in mediation effect estimation through theoretical analysis and empirical evaluations. Specifically, we have proved the identifiability of the joint distribution learned by IMAVAE and demonstrated the disentanglement of mediators in control and treatment groups. Overall, our proposed method offers a promising avenue for CMA in settings with much more complex data, where traditional methods may struggle to provide accurate estimates. 

\bibliographystyle{unsrtnat}
\bibliography{references.bib}

\newpage
\appendix

\section{Detailed Proof of Identifiability}
\renewcommand\thefigure{\thesection\arabic{figure}}
\setcounter{figure}{0}
\label{appx:A}
The proof of Theorem \ref{theo} closely resembles the proof presented in \cite{khemakhem2020variational}, which consists of 3 main steps.

\paragraph{Step 1}
The first step is to transform the equality of observed data distributions into equality of noise-free distributions using a convolutional trick based on the $1^{st}$ assumption in Theorem \ref{theo}. Similar to \cite{khemakhem2020variational}, we introduce the volume of a matrix denoted by $\text{vol}\, A$ as the product of singular values of $A$. When $A$ is full column rank, $\text{vol}\, A = \sqrt{\text{det}\, A^{\text{T}}A}$, and when $A$ is invertible, $\text{vol}\, A = |\text{det}\, A|$. We use this matrix volume as a replacement for the absolute determinant of Jacobian \cite{ben1999change} in the change of variables formula, which is most useful when the Jacobian is a rectangular matrix $(d < D)$. Suppose we have two sets of parameters $(\textbf{f}, \textbf{h}, \textbf{S}, \boldsymbol{\lambda})$ and $(\tilde{\textbf{f}}, \tilde{\textbf{h}}, \tilde{\textbf{S}}, \tilde{\boldsymbol{\lambda}})$ such that $p_{\textbf{f}, \textbf{h}, \textbf{S}, \boldsymbol{\lambda}}(\boldsymbol{x},y|\boldsymbol{u}) = p_{\tilde{\textbf{f}}, \tilde{\textbf{h}}, \tilde{\textbf{S}}, \tilde{\boldsymbol{\lambda}}}(\boldsymbol{x}, y|\boldsymbol{u})$.
% \begin{comment}
% \begin{align}
% \int_{\mathcal{Z}} p_{\textbf{S},\boldsymbol{\lambda}}(\boldsymbol{z}|\boldsymbol{u}) p_{\textbf{f}}(\boldsymbol{x}|\boldsymbol{z}) p_{\textbf{h}}(y|\boldsymbol{z},\boldsymbol{u}) d\boldsymbol{z} &= \int_{\mathcal{Z}} p_{\tilde{\textbf{S}},\tilde{\boldsymbol{\lambda}}}(\boldsymbol{z}|\boldsymbol{u}) p_{\tilde{\textbf{f}}}(\boldsymbol{x}|\boldsymbol{z}) p_{\tilde{\textbf{h}}}(y|\boldsymbol{z},\boldsymbol{u}) d\boldsymbol{z}, \\
% \int_{\mathcal{Z}} p_{\textbf{S},\boldsymbol{\lambda}}(\boldsymbol{z}|\boldsymbol{u}) p_{\boldsymbol{\epsilon}}(\boldsymbol{x} - \textbf{f}(\boldsymbol{z})) p_{\boldsymbol{\xi}}(y - \textbf{h}(\boldsymbol{z}, \boldsymbol{u})) d\boldsymbol{z} &= \int_{\mathcal{Z}} p_{\tilde{\textbf{S}},\tilde{\boldsymbol{\lambda}}}(\boldsymbol{z}|\boldsymbol{u}) p_{\boldsymbol{\epsilon}}(\boldsymbol{x} - \tilde{\textbf{f}}(\boldsymbol{z})) p_{\boldsymbol{\xi}}(y - \tilde{\textbf{h}}(\boldsymbol{z},\boldsymbol{u})) d\boldsymbol{z}.
% \end{align}
% \end{comment}
Recall that $\boldsymbol{\psi} = (\textbf{f}, \textbf{h}): \mathcal{Z} \times \mathcal{U} \rightarrow \mathcal{X} \times \mathcal{Y}$ and define the concatenated vector of $\boldsymbol{x}$ and $y$ as $\boldsymbol{v}$. We have:
\begin{align}
\int_{\mathcal{Z}} p_{\textbf{S},\boldsymbol{\lambda}}(\boldsymbol{z}|\boldsymbol{u}) p_{\boldsymbol{\psi}}(\boldsymbol{v}|\boldsymbol{z}, \boldsymbol{u}) d\boldsymbol{z} &= \int_{\mathcal{Z}} p_{\tilde{\textbf{S}},\tilde{\boldsymbol{\lambda}}}(\boldsymbol{z}|\boldsymbol{u}) p_{\tilde{\boldsymbol{\psi}}}(\boldsymbol{v}|\boldsymbol{z}, \boldsymbol{u}) d\boldsymbol{z}, \\
\int_{\mathcal{Z}} p_{\textbf{S},\boldsymbol{\lambda}}(\boldsymbol{z}|\boldsymbol{u}) p_{\boldsymbol{\epsilon},\boldsymbol{\xi}}(\boldsymbol{v} - \boldsymbol{\psi}(\boldsymbol{z},\boldsymbol{u})) d\boldsymbol{z} &= \int_{\mathcal{Z}} p_{\tilde{\textbf{S}},\tilde{\boldsymbol{\lambda}}}(\boldsymbol{z}|\boldsymbol{u}) p_{\boldsymbol{\epsilon},\boldsymbol{\xi}}(\boldsymbol{v} - \tilde{\boldsymbol{\psi}}(\boldsymbol{z},\boldsymbol{u}))  d\boldsymbol{z}.
\end{align}
Next, we apply change of variables $\bar{\boldsymbol{v}} = \boldsymbol{\psi}(\boldsymbol{z}, \boldsymbol{u})$ on the left hand side (LHS) and $\bar{\boldsymbol{v}} = \tilde{\boldsymbol{\psi}}(\boldsymbol{z}, \boldsymbol{u})$ on the right hand side (RHS):
\begin{equation}
\begin{aligned}
&\int_{\mathcal{X} \times \mathcal{Y}} p_{\textbf{S},\boldsymbol{\lambda}}(\boldsymbol{\psi}^{-1}_{|\boldsymbol{z}}(\boldsymbol{v})|\boldsymbol{u}) p_{\boldsymbol{\epsilon},\boldsymbol{\xi}}(\boldsymbol{v} - \bar{\boldsymbol{v}}) \text{vol}\, J_{\boldsymbol{\psi}^{-1}}(\bar{\boldsymbol{v}}) d \bar{\boldsymbol{v}} \\
&= \int_{\mathcal{X} \times \mathcal{Y}}  p_{\tilde{\textbf{S}},\tilde{\boldsymbol{\lambda}}}(\tilde{\boldsymbol{\psi}}^{-1}_{|\boldsymbol{z}}(\boldsymbol{v})|\boldsymbol{u}) p_{\boldsymbol{\epsilon},\boldsymbol{\xi}}(\boldsymbol{v} - \bar{\boldsymbol{v}}) \text{vol}\, J_{\tilde{\boldsymbol{\psi}}^{-1}}(\bar{\boldsymbol{v}}) d \bar{\boldsymbol{v}} , \label{eq:13}
\end{aligned}
\end{equation}
where $J$ denotes the Jacobian and recall that $\boldsymbol{\psi}^{-1}_{|\boldsymbol{z}}$ is the projection of $\boldsymbol{\psi}^{-1}$ on $\mathcal{Z}$. Next, we introduce $\tilde{p}_{\textbf{S},\boldsymbol{\lambda},\boldsymbol{\psi},\boldsymbol{u}}(\boldsymbol{v}) = p_{\textbf{S},\boldsymbol{\lambda}}(\boldsymbol{\psi}^{-1}_{|\boldsymbol{z}}(\boldsymbol{v})|\boldsymbol{u}) \text{vol}\, J_{\boldsymbol{\psi}^{-1}}(\boldsymbol{v}) \mathbb{I}_{\mathcal{X} \times \mathcal{Y}}(\boldsymbol{v})$ on the LHS and similarly on the RHS, then, following \cite{khemakhem2020variational}, Equation \ref{eq:13} reduces to:
\begin{align}
\int_{\mathcal{X} \times \mathcal{Y}} \tilde{p}_{\textbf{S},\boldsymbol{\lambda},\boldsymbol{\psi},\boldsymbol{u}}(\bar{\boldsymbol{v}}) p_{\boldsymbol{\epsilon},\boldsymbol{\xi}}(\boldsymbol{v} - \bar{\boldsymbol{v}}) d\bar{\boldsymbol{v}} &= \int_{\mathcal{X} \times \mathcal{Y}} \tilde{p}_{\tilde{\textbf{S}},\tilde{\boldsymbol{\lambda}},\tilde{\boldsymbol{\psi}},\boldsymbol{u}}(\bar{\boldsymbol{v}}) p_{\boldsymbol{\epsilon},\boldsymbol{\xi}}(\boldsymbol{v} - \bar{\boldsymbol{v}}) d\bar{\boldsymbol{v}} , \\
(\tilde{p}_{\textbf{S},\boldsymbol{\lambda},\boldsymbol{\psi},\boldsymbol{u}} * p_{\boldsymbol{\epsilon},\boldsymbol{\xi}})(\boldsymbol{v}) &= (\tilde{p}_{\tilde{\textbf{S}},\tilde{\boldsymbol{\lambda}},\tilde{\boldsymbol{\psi}},\boldsymbol{u}} * p_{\boldsymbol{\epsilon},\boldsymbol{\xi}})(\boldsymbol{v}) , \label{eq:15} \\
F[\tilde{p}_{\textbf{S},\boldsymbol{\lambda},\boldsymbol{\psi},\boldsymbol{u}}](\omega) \varphi_{\boldsymbol{\epsilon},\boldsymbol{\xi}}(\omega) &= F[\tilde{p}_{\tilde{\textbf{S}},\tilde{\boldsymbol{\lambda}},\tilde{\boldsymbol{\psi}},\boldsymbol{u}}](\omega) \varphi_{\boldsymbol{\epsilon},\boldsymbol{\xi}}(\omega) , \label{eq:16} \\
F[\tilde{p}_{\textbf{S},\boldsymbol{\lambda},\boldsymbol{\psi},\boldsymbol{u}}](\omega) &= F[\tilde{p}_{\tilde{\textbf{S}},\tilde{\boldsymbol{\lambda}},\tilde{\boldsymbol{\psi}},\boldsymbol{u}}](\omega) , \label{eq:17} \\
\tilde{p}_{\textbf{S},\boldsymbol{\lambda},\boldsymbol{\psi},\boldsymbol{u}}(\boldsymbol{v}) &= \tilde{p}_{\tilde{\textbf{S}},\tilde{\boldsymbol{\lambda}},\tilde{\boldsymbol{\psi}},\boldsymbol{u}}(\boldsymbol{v}) , \label{eq:18}
\end{align}
where we use $*$ for the convolutional operator in Equation \ref{eq:15}, we use $F[\cdot]$ to designate the Fourier transform in Equation \ref{eq:16}, $\varphi_{\boldsymbol{\epsilon},\boldsymbol{\xi}} = F[p_{\boldsymbol{\epsilon},\boldsymbol{\xi}}]$ according to the definition of the characteristic function, and we drop $\varphi_{\boldsymbol{\epsilon},\boldsymbol{\xi}}(\omega)$ from both sides in Equation \ref{eq:17} as it is non-zero almost everywhere by the $1^{st}$ assumption in Theorem \ref{theo}. 

Equation \ref{eq:18} is valid for all $(\boldsymbol{v}, \boldsymbol{u}) \in \mathcal{X} \times \mathcal{Y} \times \mathcal{U}$. It indicates that for the observed data distributions to be the same, the noise-free distributions have to be the same.

\paragraph{Step 2}
The second step in \cite{khemakhem2020variational} is about removing all terms that are either a function of observations $\boldsymbol{x}$ and $y$ or auxiliary variables $\boldsymbol{u}$, which is done by introducing the points provided by the $4^{th}$ assumption in Theorem \ref{theo}, and using $\boldsymbol{u}_0$ as a ``pivot''. Specifically, by taking the logarithm on both sides of Equation \ref{eq:18} and plugging in Equation \ref{eq:6} for $p_{\textbf{S},\boldsymbol{\lambda}}$, we get:
{\allowdisplaybreaks
\begin{equation}
\begin{split}
&\log \text{vol}\, J_{\boldsymbol{\psi}^{-1}}(\boldsymbol{v}) + \sum_{i=1}^{d} \left( \log Q_i (\psi^{-1}_{i\,|\boldsymbol{z}}(\boldsymbol{v})) - \log C_i(\boldsymbol{u}) + \sum_{j=1}^{k} S_{i,j}(\psi^{-1}_{i\,|\boldsymbol{z}}(\boldsymbol{v})) \lambda_{i,j}(\boldsymbol{u}) \right) \\
&= \log \text{vol}\, J_{\tilde{\boldsymbol{\psi}}^{-1}}(\boldsymbol{v}) + \sum_{i=1}^{d} \left( \log \tilde{Q}_i (\tilde{\psi}^{-1}_{i\,|\boldsymbol{z}}(\boldsymbol{v})) - \log \tilde{C}_i(\boldsymbol{u}) + \sum_{j=1}^{k} \tilde{S}_{i,j}(\tilde{\psi}^{-1}_{i\,|\boldsymbol{z}}(\boldsymbol{v})) \tilde{\lambda}_{i,j}(\boldsymbol{u}) \right) . \label{eq:19}
\end{split}
\end{equation}}
Let $\boldsymbol{u}_0, ..., \boldsymbol{u}_{dk}$ be the points provided by the $4^{th}$ assumption of Theorem \ref{theo}, and define $\bar{\boldsymbol{\lambda}}(\boldsymbol{u}) \coloneqq \boldsymbol{\lambda}(\boldsymbol{u}) - \boldsymbol{\lambda}(\boldsymbol{u}_0)$. Then for $l = 1, ..., dk$, we plug each of those $\boldsymbol{u}_l$ in Equation \ref{eq:19} to obtain $dk + 1$ equations and subtract the first equation for $\boldsymbol{u}_0$ from the remaining $dk$ equations to get:
\begin{equation}
\langle \textbf{S}(\boldsymbol{\psi}^{-1}_{|\boldsymbol{z}}(\boldsymbol{v})), \bar{\boldsymbol{\lambda}}(\boldsymbol{u}_l) \rangle + \sum_{i=1}^{d} \log \frac{C_i(\boldsymbol{u}_0)}{C_i(\boldsymbol{u}_l)} = \langle \tilde{\textbf{S}}(\tilde{\boldsymbol{\psi}}^{-1}_{|\boldsymbol{z}}(\boldsymbol{v})), \bar{\tilde{\boldsymbol{\lambda}}}(\boldsymbol{u}_l) \rangle + \sum_{i=1}^{d} \log \frac{\tilde{C}_i(\boldsymbol{u}_0)}{\tilde{C}_i(\boldsymbol{u}_l)} , \label{eq:20}
\end{equation}
where $\langle \cdot , \cdot \rangle$ denotes inner product. Let $L$ be the matrix defined in the $4^{th}$ assumption of Theorem \ref{theo}, and $\tilde{L}$ similarly for $\tilde{\boldsymbol{\lambda}}$ ($\tilde{L}$ is not necessarily invertible). Define $b_l = \sum_{i=1}^{d} \log \frac{\tilde{C}_i(\boldsymbol{u}_0) C_i(\boldsymbol{u}_l)}{C_i(\boldsymbol{u}_0) \tilde{C}_i(\boldsymbol{u}_l)}$ and $\textbf{b}$ the vector of all $b_l$ for $l = 1, ..., dk$. Expressing Equation \ref{eq:20} in matrix form, we get:
\begin{equation}
L^{\text{T}} \textbf{S}(\boldsymbol{\psi}^{-1}_{|\boldsymbol{z}}(\boldsymbol{v})) = \tilde{L}^{\text{T}} \tilde{\textbf{S}}(\tilde{\boldsymbol{\psi}}^{-1}_{|\boldsymbol{z}}(\boldsymbol{v})) + \textbf{b}. \label{eq:21}
\end{equation}
Multiplying both sides of Equation \ref{eq:21} by the transpose of the inverse of $L^{\text{T}}$, we get:
\begin{equation}
\textbf{S}(\boldsymbol{\psi}^{-1}_{|\boldsymbol{z}}(\boldsymbol{v})) = A \tilde{\textbf{S}}(\tilde{\boldsymbol{\psi}}^{-1}_{|\boldsymbol{z}}(\boldsymbol{v})) + \textbf{c}, \label{eq:22}
\end{equation}
where $A = L^{-\text{T}} \tilde{L}$ and $\textbf{c} = L^{-\text{T}} \textbf{b}$. Note that here $\boldsymbol{\psi}^{-1}_{|\boldsymbol{z}}(\boldsymbol{v})$ can be referred to either the projection from $\textbf{f}$ or from $\textbf{h}$ as we map the same $Z$ into $\hat{X}$ and $Y$ through $p_{\textbf{f}}$ and $p_{\textbf{h}}$, respectively.

\paragraph{Step 3}
In the last step, Khemakhem et al. \cite{khemakhem2020variational} show that the linear transformation $A$ is invertible for both $k = 1$ and $k > 1$, resulting in an equivalence relation in the iVAE framework. This is mainly based on the $3^{rd}$ assumption in Theorem \ref{theo} which guarantees the existence of the $dk \times d$ Jacobian matrix of $\textbf{S}$ with rank $d$. This also implies that the Jacobian of $\tilde{\textbf{S}} \circ \tilde{\boldsymbol{\psi}}^{-1}_{|\boldsymbol{z}}$ exists and is of rank $d$ and so is $A$. We argue that the proof of invertibility in our framework follows the same line of reasoning as that of the iVAE.

Therefore, with Equation \ref{eq:22} and the invertibility of $A$, we prove that the parameters $(\textbf{f}, \textbf{h}, \textbf{S}, \boldsymbol{\lambda})$ are $\sim_{A}$-identifiable.

\section{Details of Experimental Setup}
\renewcommand\thefigure{\thesection\arabic{figure}}
\setcounter{figure}{0}
\label{appx:B}
\paragraph{Model Configuration}
In all experiments, the encoder $q_{\boldsymbol{\phi}}(\boldsymbol{z}|\boldsymbol{x},\boldsymbol{u})$, decoder $p_{\textbf{f}}(\boldsymbol{x}|\boldsymbol{z})$, and prior $p_{\textbf{S},\boldsymbol{\lambda}}(\boldsymbol{z}|\boldsymbol{u})$ of IMAVAE are configured as multivariate normal distributions whose mean and covariance are parameterized as a function of the conditioned variables using a feed-forward neural network. As for the predictor $g_{\boldsymbol{\gamma}}(\boldsymbol{z},\boldsymbol{u})$, it is implemented as a simple linear regression model. It is worth noting that while alternative stochastic models can also be employed to replace $g_{\boldsymbol{\gamma}}$, our experiments demonstrate that the linear regression model already achieves superior performance.  See the included code link for details on reproduction of all experiments.

\paragraph{Training Details}
When training IMAVAE to minimize the objective in Equation \ref{eq:8}, we use the Adam optimizer and adopt parameter annealing so that the KL divergence will gradually dominate the reconstruction error in ELBO.

\paragraph{Computing Resources}
The experiments conducted on the synthetic data and the Jobs II data in Sections \ref{sec:5.1} and \ref{sec:5.3} are of a relatively small scale and can be executed locally. The experiment on the electrophysiological data is performed on a computer cluster equipped with a GeForce RTX 2080 Ti GPU. 

\paragraph{Data Availability}

Multi-region local field potential recordings during a tail-suspension test is an experiment comparing the electrical neural activity and behaviors of Wildtype and Clock-$\Delta$19 genotypes of mice in the tail-suspension test. This dataset is available to download at \url{https://research.repository.duke.edu/concern/datasets/zc77sr31x?locale=en} for free under a Creative Commons BY-NC Attribution-NonCommercial 4.0 International license. 

Jobs II is a randomized field experiment that investigates the efficacy of a job training intervention on unemployed workers, which can be downloaded from the R package "mediation" or the following URL: \url{https://r-data.pmagunia.com/dataset/r-dataset-package-mediation-jobs}.

\section{Synthetic Data Generation}
\renewcommand\thefigure{\thesection\arabic{figure}}
\setcounter{figure}{0}
\label{appx:C}
The synthetic data contains $N = 6000$ data points and is generated according to the causal graphs shown in Figure \ref{fig:1}. Specifically, for case (a) without observed covariates, we generate $t_i$, $\boldsymbol{z}_i$, $\boldsymbol{x}_i$, and $y_i$ for $i = 1, ..., N$ as follows:
\begin{align*}
t_i &\sim \text{Bernoulli}(p), \\
\boldsymbol{z}_i &\sim \mathcal{N}(\boldsymbol{0}, \sigma_z^2 \textbf{I}_d) + c \mathbb{I}(t_i = 1) \textbf{1}_d, \\
\boldsymbol{x}_i &= f(\boldsymbol{z}_i) + \boldsymbol{\epsilon}_x, \\
y_i &= \mu_t t_i + \boldsymbol{\mu}_z^{\text{T}} \boldsymbol{z}_i + \epsilon_y,
\end{align*}
where $p$ is a probability parameter, $\sigma_z^2$ is a variance parameter, $\textbf{I}_d$ is a $d \times d$ identity matrix, $\mathbb{I}(\cdot)$ denotes the indicator function, $\textbf{1}_d$ is a $d$-dimensional vector of all ones, $f: \mathbb{R}^d \rightarrow \mathbb{R}^D$ is a nonlinear function modeled by an un-trained neural network, $\boldsymbol{\epsilon}_x \in \mathbb{R}^D$ and $\epsilon_y \in \mathbb{R}$ are Gaussian noise terms, and $c$, $\mu_t$, and $\boldsymbol{\mu}_z$ are coefficient constants/vector. For case (b) with observed covariates, we generate $t_i$, $\boldsymbol{z}_i$, $\boldsymbol{x}_i$, $\boldsymbol{w}_i$, and $y_i$ for $i = 1, ..., N$ as follows:
\begin{align*}
\boldsymbol{w}_i &\sim \mathcal{N}(\boldsymbol{0}, \sigma_w^2 \textbf{I}_m), \\
t_i &\sim \text{Bernoulli}(\text{sigmoid}(\boldsymbol{\mu}_s^{\text{T}} \boldsymbol{w}_i)), \\
\boldsymbol{z} &\sim \mathcal{N} (\boldsymbol{0}, \sigma_z^2 \textbf{I}_d) + c_1 \mathbb{I}(t_i = 1)\textbf{1}_d + c_2 f_1(\boldsymbol{w}_i), \\
\boldsymbol{x}_i &= f_2 (\boldsymbol{z}_i) + \boldsymbol{\epsilon}_x, \\
y_i &= \mu_t t_i + \boldsymbol{\mu}_z^{\text{T}} \boldsymbol{z}_i + \boldsymbol{\mu}_w^{\text{T}} \boldsymbol{w}_i + \epsilon_y,
\end{align*}
where $\sigma_w^2$, $\sigma_z^2$ are variance parameters, $\textbf{I}_m, \textbf{I}_d$ are identity matrices with dimension $m \times m$ and $d \times d$, respectively, $\mathbb{I}(\cdot)$ denotes the indicator function, $\textbf{1}_d$ is a $d$-dimensional vector of all ones, $f_1: \mathbb{R}^m \rightarrow \mathbb{R}^d, f_2: \mathbb{R}^d \rightarrow \mathbb{R}^D$ are nonlinear functions modeled by un-trained neural networks, $\boldsymbol{\epsilon}_x \in \mathbb{R}^D$ and $\epsilon_y \in \mathbb{R}$ are Gaussian noise terms, and $c_1, c_2, \mu_t, \boldsymbol{\mu}_s, \boldsymbol{\mu}_z, \boldsymbol{\mu}_w$ are coefficient constants/vectors.

\section{Post-processing for Tail Suspension Test Data}
\renewcommand\thefigure{\thesection\arabic{figure}}
\setcounter{figure}{0}
\label{appx:D}
As elaborated in Section \ref{sec:5.2}, we have the LFP signals from 26 mice. The full dataset contains a total of $N = 43882$ data points. The observed feature, i.e. $\boldsymbol{x}_i$, represents the power spectral densities of the LFPs recorded at 11 brain regions. These densities are evaluated from $1$ to $56 Hz$, resulting in a total of $616$ attributes, i.e. $\boldsymbol{x}_i \in \mathbb{R}^{616}$. We also have treatment assignment $t_i$ indicating whether the mouse corresponding to the $i^{th}$ data point is recorded during an open field exploration $(t_i = 0)$ or a tail suspension test $(t_i = 1)$. 

To generate the semi-synthetic dataset, we do some post-processing on these data. For case (a) without observed covariates, we first apply PCA to map $\boldsymbol{x}_i$ into a lower-dimensional representation $\boldsymbol{s}_i$. Then the true mediator $\boldsymbol{z}_i$ is generated by $\boldsymbol{z}_i = \boldsymbol{s}_i + \mathbb{I}(t_i = 1)$ where $\mathbb{I}(\cdot)$ denotes the indicator function. The final outcome $y_i$ is modeled by $y_i = \mu_t t_i + \boldsymbol{\mu}_z \boldsymbol{z}_i + \epsilon_y$ where $\mu_t, \boldsymbol{\mu}_z$ are coefficient constant/vector and $\epsilon_y$ is a Gaussian noise term. For case (b), we use the genotype of the mouse $w_i \in \{0,1\}$ as an observed covariate where $w_i = 0$ denotes wild type and $w_i = 1$ denotes Clock$\Delta 19$ mutation. The true mediator $\boldsymbol{z}_i$ is then generated by $\boldsymbol{z}_i = \boldsymbol{s}_i + \mathbb{I}(t_i = 1) + f(\boldsymbol{w}_i)$ where $f: \mathbb{R} \rightarrow \mathbb{R}^d$ is a nonlinear function modeled by a neural network. The final outcome $y_i$ is modeled by $y_i = \mu_t t_i + \boldsymbol{\mu}_z \boldsymbol{z}_i + \mu_w w_i + \epsilon_y$ where $\mu_t, \mu_w, \boldsymbol{\mu}_z$ are coefficient constants/vector and $\epsilon_y$ is a Gaussian noise term.

\section{Simulation Procedure for Jobs II Data}
\renewcommand\thefigure{\thesection\arabic{figure}}
\setcounter{figure}{0}
\label{appx:E}
To achieve \emph{zero} direct, mediation, and total effects, we adopt the following simulation procedure, similar to \cite{cheng2022causal,huber2016finite}, on the Jobs II dataset. Note that all attributes other than $T, Z, Y$ are treated as observed covariates $W$ in this analysis. $X$ in our causal graphs does not exist in this dataset.
\begin{enumerate}[leftmargin=*]
\item Estimate probit specifications in which we regress (a) $T$ on $W$ to get an estimated probit coefficient $\hat{\beta}_{\text{pop}}$ and (b) $Z$ on $T$ and $W$ to get estimated probit coefficients $\hat{\gamma}_{\text{pop}}$ and $\hat{\delta}_{\text{pop}}$, respectively.
\item Apply the indicator function to $Z$ so that the mediator becomes a binary variable $Z \coloneqq \mathbb{I}(Z \geq 3)$.
\item Discard all samples with either $T = 0$ or $Z = 0$, resulting in a dataset with all non-mediated and non-treated samples.
\item Draw independent Monte Carlo samples (500 and 1000 samples for Tables \ref{tab:4} and \ref{tab:5}, respectively) with replacement $(T', Z', W', Y')$ from the resulting dataset.
\item Simulate the (pseudo-)treatment and (pseudo-)mediator using the following formula:
\begin{align*}
T' &\coloneqq \mathbb{I}(W' \hat{\beta}_{\text{pop}} + U > 0), \\
M' &\coloneqq \eta (T' \hat{\gamma}_{\text{pop}} + W' \hat{\delta}_{\text{pop}}) + \alpha + V,
\end{align*}
where $U \sim \mathcal{N}(0,1)$, $V \sim \mathcal{N}(0,1)$, $\eta$ is a simulation parameter which stands for the magnitude of selection into the mediator, and we manually set $\alpha$ such that either $10\%$ or $50\%$ of the samples are mediated (i.e. $M' \geq 3$). Note that here the (pseudo-)mediator $M'$ is continuous after simulation.
\end{enumerate}
With this simulation design, the ground truth of the direct, mediated, and total effects are all \emph{zero}.

%%%%%%%%%%%%%%%%%%%%%%%%%%%%%%%%%%%%%%%%%%%%%%%%%%%%%%%%%%%%

\end{document}